\documentclass[10pt,twocolumn,letterpaper]{article}

\usepackage{iccv}
\usepackage{times}
\usepackage{epsfig}
\usepackage{graphicx}
\usepackage{amsmath}
\usepackage{amssymb}

\usepackage[breaklinks=true,bookmarks=false]{hyperref}

\iccvfinalcopy % *** Uncomment this line for the final submission

\def\iccvPaperID{0000} % *** Enter the ICCV Paper ID here
\ificcvfinal\fi

\begin{document}

%%%%%%%%% TITLE
\title{On the Robustness of Vision Transformers to Adversarial Examples}

\author{Kaleel Mahmood\\
Department of Computer Science and Engineering\\
University of Connecticut, CT, 06269, USA 
{\tt\small kaleel.mahmood@uconn.edu}

\and
Rigel Mahmood\\
Department of Computer Science and Engineering\\
University of Connecticut, CT, 06269, USA 
\and
Marten Van Dijk\\
CWI, Amsterdam\\
The Netherlands 
}

\maketitle
\ificcvfinal\thispagestyle{empty}\fi

%%%%%%%%% ABSTRACT
\begin{abstract}
Recent advances in attention-based networks have shown that Vision Transformers can achieve state-of-the-art or near state-of-the-art results on many image classification tasks. This puts transformers in the unique position of being a promising alternative to traditional convolutional neural networks (CNNs). While CNNs have been carefully studied with respect to adversarial attacks, the same cannot be said of Vision Transformers. In this paper, we study the robustness of Vision Transformers to adversarial examples. Our analyses of transformer security is divided into three parts. First, we test the transformer under standard white-box and black-box attacks. Second, we study the transferability of adversarial examples between CNNs and transformers. We show that adversarial examples do not readily transfer between CNNs and transformers. Based on this finding, we analyze the security of a simple ensemble defense of CNNs and transformers. By creating a new attack, the self-attention blended gradient attack, we show that such an ensemble is not secure under a white-box adversary. However, under a black-box adversary, we show that an ensemble can achieve unprecedented robustness without sacrificing clean accuracy. Our analysis for this work is done using six types of white-box attacks and two types of black-box attacks. Our study encompasses multiple Vision Transformers, Big Transfer Models and CNN architectures trained on CIFAR-10, CIFAR-100 and ImageNet.
\end{abstract}
\section{Introduction}
\label{sec:intro}

For vision tasks, convolutional neural networks (CNNs)~\cite{LeCunCNN} are the de facto architecture~\cite{xie2020selftraining, Kolesnikov_2020}. On the other hand, in natural language processing (NLP), attention-based transformers are one of the most commonly used models~\cite{vaswani2017attention}. %In particular, self-attention based transformers have been widely adopted~\cite{BERT}. The training strategy for self-attention transformers is to first train on a complex, larger dataset before being fine-tuned on a specific, smaller dataset~\cite{BERT}. 
Based on the success of transformers in NLP, various works have attempted to apply self-attention (both with and without CNNs) to image processing tasks ~\cite{Carion_2020, wang2020axial}. In particular, in~\cite{dosovitskiy2021an}, the training of self-attention transformers is achieved by processing the image in patches. The training in~\cite{dosovitskiy2021an} is unique in that the transformer is first trained on the dataset ImageNet-21K (or JFT) before training on a smaller dataset, to achieve near state-of-the-art results on ImageNet, CIFAR-10 and CIFAR-100. These types of transformers are referred to as Vision Transformers (ViT)~\cite{dosovitskiy2021an}. It is important to note that the same kind of training regime can be applied to CNNs. In~\cite{Kolesnikov_2020}, they also propose training on a large dataset (ImageNet-21K or JFT) and fine tuning on a smaller dataset. Using this approach, CNNs are also able to achieve state-of-the-art results on ImageNet, CIFAR-10 and CIFAR-100. CNNs trained in this manner are referred to as Big Transfer Models (BiT-M)~\cite{Kolesnikov_2020}.

While CNNs are popular for vision tasks, they are not without deficiencies. It has been widely documented that CNNs are vulnerable to adversarial examples~\cite{szegedy2014intriguing,goodfellow2014explaining}. Adversarial examples are benign input images to which small perturbations are added. This perturbation causes the CNN to misclassify the image with high confidence. Broadly speaking, an attacker creates an adversarial example using one of two threat models. Under a white-box adversary~\cite{carlini2019evaluating}, the attacker has access to the CNN’s parameters (architecture and trained weights). The adversary can directly obtain gradient information from the model to create an adversarial example. The other type of threat is a black-box adversary. In this scenario, the attacker does not know the CNN’s parameters or architecture but can repeatedly query the CNN, or build their own synthetic CNN to estimate gradient information and generate adversarial examples. 

It has also been shown that adversarial examples generated using CNNs exhibit transferability~\cite{papernot2016transferability,liu2016delving,papernot2017practical}. Here, transferability refers to the fact that adversarial examples crafted to fool one CNN are often misclassified by other CNNs as well. Overall, CNNs have an expansive body of literature related to adversarial attacks~\cite{carlini2017towards, MIMdong2018boosting, croce2020reliable} and defenses~\cite{madry2018towards, carlini2019evaluating,tramer2020adaptive}. In contrast, Vision Transformers have not been closely studied in the adversarial context. In this work, we investigate how the advent of Vision Transformers advance the field of adversarial machine learning. Here we specifically focus on image based adversarial attacks. Our paper is organized as follows: In section~\ref{sec:related} we first discuss some related NLP work. We then break our analysis of Vision Transformers into several related questions:

\textit{Do Vision Transformers provide any improvement in security over CNNs under a white-box adversary?} We explore this question in section~\ref{sec:whitebox} by attacking Vision Transformers, Big Transfer Models and conventional CNNs (ResNets) with six standard white-box adversarial machine learning attacks. We show that under a white-box adversary, Vision Transformers are just as vulnerable (insecure) as other models. In section~\ref{sec:transfer}, we further delve into white-box attacks and ask: \textit{How transferable are adversarial examples between Visions Transformers and other models?} We perform a transferability study with eight CIFAR-10 and CIFAR-100 models (this includes four Vision Transformers, two Big Transfer Models, and two ResNets). We also study the transferability of ImageNet Vision Transformers using seven models (three Vision Transformers, two Big Transfer Models, and two ResNets). From our experiments we observe an interesting phenomenon. The transferability between Vision Transformers and other non-transformer models is unexpectedly low.

\textit{How can the transferability phenomena be leveraged to provide security?} This is the topic of our final question in sections~\ref{sec:ensemble} and ~\ref{sec:bb}. We further break this question down into white-box and black-box analyses. First, we consider a white-box adversary. We develop a new white-box attack called the Self-Attention blended Gradient Attack (SAGA). Using SAGA, we show it is not possible to leverage the transferability phenomena to achieve white-box security. However, achieving black-box security is still possible. To demonstrate this, we consider a black-box attacker that can leverage transfer style~\cite{papernot2017practical} and query-based attacks~\cite{chen2020rays}. We show under this threat model, a simple ensemble of Vision Transformers and Big Transfer Models can achieve an unprecedent level of robustness, without sacrificing clean accuracy. Finally, in section~\ref{sec:conclusion}, we offer concluding remarks. 

\section{Related Work}
\label{sec:related}
The transformer has been well studied from an adversarial perspective for NLP applications e.g.,~\cite{hsieh2019robustness, shi2020robustness, hendrycks-etal-2020-pretrained, hao2020self}. The work in~\cite{hsieh2019robustness} analyzes two popular self-attentive architectures: (a) Transformer for neural machine translation, and (b) BERT for sentiment and entailment classification, and proposes algorithms to generate more natural adversarial examples that preserve the semantics. Theoretical explanations are also provided in~\cite{hsieh2019robustness} to support the claim that self-attentive structures are more robust to small adversarial perturbations in NLP as compared to LSTM based architectures. The work in~\cite{shi2020robustness} analyzes the complex relationship between self-attention layers including cross-non-linearity and cross-position, and develops a robustness verification algorithm for Transformers. The authors do not use large-scale pre-trained models such as BERT because they are too challenging to be tightly verified
with their approach. The work in~\cite{hendrycks-etal-2020-pretrained} studies large pre-trained Transformer models in NLP such as BERT. One of the conjectures drawn by the authors of~\cite{hendrycks-etal-2020-pretrained} is that since Transformer models are pre-trained with large amounts of data (e.g., BERT is trained on 3 billion tokens), this may aid robustness. It is also mentioned that perhaps the self-supervised training may also contribute to this robustness. The work in~\cite{hao2020self} proposes a self-attention attribution method to interpret the information interactions inside a transformer. The authors use BERT as an example to conduct experiments to identify the important attention heads, and extract the most salient dependencies in each layer to construct an attribution tree. This information is used to extract adversarial patterns to implement non-targeted attacks towards BERT.

Thus, as stated above a good body of work has been devoted to the adversarial exploration of the Transformer for NLP applications. To our best knowledge, we are the first to provide an in-depth analysis of the adversarial properties of a Transformer from a vision perspective.

\section{White-Box Attacks on Vision Transformers}
\label{sec:whitebox}

 \textit{Do Vision Transformers provide any improvement in security  over  CNNs  under  a  white-box  adversary?} We experimentally analyze Vision Transformers to answer this question. It may seem unorthodox to start with experiments. However, the most expedient way to directly determine the security of the transformer is through attacks and analyses of those attacks. We start with a white-box adversary because it represents the strongest possible adversary.

\subsection{Adversarial Model, Considered Classifiers  and White-Box Attack Selection}
\label{subsection:wsetup}

\textbf{Adversarial Model:} In this section, our adversary has knowledge of the model architecture and trained parameters of the model. We assume the adversary can perturb the original input $x$ to create $x_{adv}$ within a certain amount $\epsilon$ according to $\|x-x_{adv}\|_{\infty} \leq \epsilon$. For CIFAR-10 and CIFAR-100, the $\epsilon=0.031$ and for ImageNet $\epsilon=0.062$, where $x$ is an $n \times m$ color image such that $x \in {[0,1]}^{n\times m \times 3}$.  The adversary succeeds if they are able to create an input $x_{adv}$ within this bound $\epsilon$ that is misclassified by the classifier (untargeted attack). When we measure security, we do so by taking a set of clean test examples that are correctly identified by the classifier. Using this set of clean examples we generate adversarial examples using one of the six attacks. We then measure what percent of examples the classifier still correctly identifies. As Vision Transformers are relatively new, we experiment with a wide range of attacks and models. Below, we list the attacks and models we use. We also give our justification for including them in this paper.

\textbf{White-Box Attacks:} We run six different types of white-box attacks on our models. We begin with one of the most basic, the Fast Gradient Sign Method (FGSM)~\cite{ FGSMpaper} as an initial test of robustness. We further build upon this by testing stronger multi-step attacks, the Momentum Iterative Method (MIM)~\cite{MIMdong2018boosting}, and Projected Gradient Descent (PGD)~\cite{PGDmadry2018}. We also test the newest iterative attack which uses a variable step size in each iteration, Auto Projected Gradient Descent (APGD)~\cite{croce2020reliable}. Aside from the previously mentioned attacks, there are two other possible attack directions. To craft an extremely small, almost imperceptible adversarial noise, the Carlini and Wagner (C \& W) attack is often of interest~\cite{carlini2017towards}. 

Lastly, it is possible for some white-box attacks to fail if gradient masking or an obfuscation of the gradient occurs~\cite{BPDApaper}. It is important to note this does not actually mean the classifier is secure, it merely means the gradient for the classifier was not estimated properly. There are attacks designed to overcome gradient masking, such as the Backward Pass Differentiable Approximation (BPDA)~\cite{BPDApaper}. We use BPDA here to ensure gradient masking is not occurring in the self-attention layers, or any other part of the Vision Transformer. Due to the limited space, we cannot give detailed descriptions of each white-box attack here. We urge interested readers to examine the supplemental material where we provide descriptions of each attack.

\textbf{Classifier Models:} When considering Vision Transformers, there are several different types of model variants. To begin, the patch size of the transformer needs to be chosen. To test different patch sizes, in our study we include both patch size 32 (ViT-B-32) and patch size 16 (ViT-B-16). The "B" in the model refers to the model complexity~\cite{dosovitskiy2021an}. "B" models contain 12 layers and "L" models contain 24 layers. Since model complexity is another factor that can affect security~\cite{PGDmadry2018}, we also test across model complexity (ViT-B-16 and ViT-L-16). It is also possible to use the self-attention layers first and then use a conventional CNN (ResNet) on top. This configuration is denoted as ViT-R50. Experimenting across patch size, model complexity and with the hybrid configuration gives us four Vision Transformer models.

For the Big Transfer Models~\cite{Kolesnikov_2020}, we vary across model complexity (BiT-M-R50 and BiT-M-R101x3). We do the same for conventional ResNets (ResNet-56 and ResNet-164~\cite{he2016identity}). Overall for CIFAR-10 and CIFAR-100, this gives us a total of 8 models to attack: ViT-B-32, ViT-B-16, ViT-L-16, ViT-R50, BiT-M-R50, BiT-M-R101x3, ResNet-56 and ResNet-164. For ImageNet, we run a slight variation of the above set, attacking 7 models: ViT-B-16, ViT-L-16 (image size 224), ViT-L-16 (image size 512), BiT-M-R50, BiT-M-R152x4, ResNet-50 and ResNet-152. For ImageNet, we mainly focus on more complex models (e.g., testing two types of ViT-L-16 instead of ViT-B-32). We do this because the more complex Vision Transformers are better indicative of state-of-the-art performance on ImageNet. We provide full descriptions of the architectures and training parameters for our models in the supplemental material.     

\begin{table*}[htbp!]
\label{table:whiteboxattacks}
\centering
\caption{White-box attacks on Vision Transformers, Big Transfer Models and ResNets. The attacks are done using the $l_{\infty}$ norm with $\epsilon=0.031$ for CIFAR-10 and $\epsilon=0.062$ for ImageNet. The white-box attack results for CIFAR-100 follow an extremely similar trend to CIFAR-10. Hence for brevity, CIFAR-100 white-box attack results are given in the supplementary material. In this Table the robust accuracy is given for each corresponding attack. The last column "Acc" refers to the clean accuracy of the model.}
{\small
\begin{tabular}{l|c|c|c|c|c|c|c|}
\cline{2-8}
 & \multicolumn{7}{c|}{CIFAR-10} \\ \cline{2-8} 
 & FGSM & PGD & BPDA & MIM & C\&W & APGD & Acc \\ \hline
\multicolumn{1}{|l|}{ViT-B-32} & 37.9\% & 1.8\% & 17.6\% & 4.4\% & 0.0\% & 0.0\% & 98.6\% \\ \hline
\multicolumn{1}{|l|}{ViT-B-16} & 39.5\% & 0.0\% & 20.3\% & 0.3\% & 0.0\% & 0.0\% & 98.9\% \\ \hline
\multicolumn{1}{|l|}{ViT-L-16} & 56.3\% & 1.2\% & 28.7\% & 5.9\% & 0.0\% & 0.0\% & 99.1\% \\ \hline
\multicolumn{1}{|l|}{ViT-R50} & 40.8\% & 0.1\% & 13.4\% & 0.2\% & 0.0\% & 0.0\% & 98.6\% \\ \hline
\multicolumn{1}{|l|}{BiT-M-R50x1} & 66.0\% & 0.0\% & 14.9\% & 0.0\% & 0.0\% & 0.0\% & 97.5\% \\ \hline
\multicolumn{1}{|l|}{BiT-M-R101x3} & 85.2\% & 0.0\% & 17.1\% & 0.0\% & 0.0\% & 0.0\% & 98.7\% \\ \hline
\multicolumn{1}{|l|}{ResNet-56} & 23.0\% & 0.0\% & 5.0\% & 0.0\% & 0.0\% & 0.0\% & 92.8\% \\ \hline
\multicolumn{1}{|l|}{ResNet-164} & 29.0\% & 0.0\% & 5.4\% & 0.0\% & 0.0\% & 0.0\% & 93.8\% \\ \hline
 & \multicolumn{7}{c|}{ImageNet} \\ \cline{2-8} 
 & FGSM & PGD & BPDA & MIM & C\&W & APGD & Acc \\ \hline
\multicolumn{1}{|l|}{ViT-B-16} & 23.1\% & 0.0\% & 7.3\% & 0.0\% & 0.0\% & 0.0\% & 80.3\% \\ \hline
\multicolumn{1}{|l|}{ViT-L-16 (224)} & 27.9\% & 0.0\% & 8.4\% & 0.0\% & 0.0\% & 0.0\% & 82.0\% \\ \hline
\multicolumn{1}{|l|}{ViT-L-16 (512)} & 29.8\% & 0.0\% & 8.4\% & 0.0\% & 0.0\% & 0.0\% & 85.4\% \\ \hline
\multicolumn{1}{|l|}{BiT-M-R50x1} & 28.7\% & 0.0\% & 3.5\% & 0.0\% & 0.0\% & 0.0\% & 79.9\% \\ \hline
\multicolumn{1}{|l|}{BiT-M-R152x4} & 60.9\% & 0.0\% & 15.2\% & 0.0\% & 0.0\% & 0.0\% & 85.3\% \\ \hline
\multicolumn{1}{|l|}{ResNet-50} & 11.8\% & 0.0\% & 1.4\% & 0.0\% & 0.0\% & 0.0\% & 74.5\% \\ \hline
\multicolumn{1}{|l|}{ResNet-152} & 18.1\% & 0.0\% & 2.7\% & 0.0\% & 0.0\% & 0.0\% & 77.0\% \\ \hline
\end{tabular}
}
\end{table*}

\subsection{White-Box Attack Analysis}

 We report the results of our six white-box attacks for CIFAR-10 and ImageNet in Table~\ref{table:whiteboxattacks}. The robust accuracy (percent of samples correctly identified by the classifier) is reported in Table~\ref{table:whiteboxattacks} using $1000$ examples for each attack. For this set of attacks, CIFAR-10 and CIFAR-100 follow extremely similar trends. As a result, for brevity, we provide our CIFAR-100 white-box attack results in the supplementary material.  

Overall, based on the results in Table~\ref{table:whiteboxattacks}, we can definitively answer the original question posed at the start of the this section. Vision Transformers do not provide any additional security over Big Transfer Models or conventional CNNs. We can clearly see this across all datasets, indicating Vision Transformers have no robustness (i.e. $0\%$) for the C$\&$W and APGD attacks. Likewise, Vision Transformers have less than $6\%$ robustness across all the datasets for the PGD and MIM attacks. While this result may seem expected, it is an important step in understanding the complete security picture of Vision Transformers. Now that we know Vision Transformers are not robust to white-box attacks, we can consider the next important question on transferability.

\section{Vision Transformers Transferability Study}
\label{sec:transfer}

\textit{How transferable are the adversarial examples created by Vision Transformers?} It was shown in Section~\ref{sec:whitebox} that white-box attacks are extremely effective at creating examples that fool Vision Transformers. We further expand on the previous analyses and now examine the \textit{transferability} of adversarial examples misclassified by Vision Transformers. Here, transferability refers to the occurrence of adversarial examples that are misclassified by multiple (i.e., more than one) classifier. The transferability of adversarial examples has been well documented for different CNN architectures. In the literature, the transferability of adversarial examples was first observed in~\cite{szegedy2014intriguing}. Consequent studies have shown the transferability of adversarial examples between CNNs on the MNIST dataset in~\cite{PapernotMG16} and on the ImageNet dataset in~\cite{LiuTransfer2017}. However, to the best of our knowledge, there have been no large-scale studies on the transferability between CNNs and Visions Transformers at this time. We provide detailed evaluation and analyses on this aspect in this section.

\subsection{Measuring Transferability}

Formally, we can define non-targeted transferability as follows: We start with a classifier $C_{i}$ and correctly identified input/label pair $(x,y)$. An attack $A_{C_{i}}$ is used to generate an adversarial example $x_{adv}$ with respect to classifier $C_{i}$:
\begin{equation}
    x_{adv}= A_{C_{i}}(x,y)
\end{equation}
The adversarial example $x_{adv}$ is then said to \textit{transfer} from classifier to $C_{i}$ to  $n-1$ other classifiers if and only if:
\begin{equation}
\forall_{j=1}^n \ \left[ \{C_{j}(x) = y\} \land \{C_{j}(x_{adv}) \neq y\} \right]
\label{eq:main}
\end{equation}
Equation~\ref{eq:main} states that each classifier $C_{j}$ must correctly classify $x$ and must misclassify $x_{adv}$. Assuming two classifiers ($n=2$) and a set of $m$ examples that are correctly classified by both, we can define the transferability from $C_{i}$ to $C_{j}$ as follows:
\begin{equation}
t_{i,j} =\frac{1}{m} \sum_{k=1}^{m}
\left\{
\begin{array}{ll}
1 & \mbox{ if } C_{j}(A_{C_{i}}(x_{k},y_{k})) \neq y_{k}, \\
0 & \mbox{ otherwise.}
\end{array}
\right.
\label{eq:transfer}
\end{equation}

A high transferability between classifiers indicates that they have a shared vulnerability to the same set of adversarial examples. On the other hand, a low transferability may indicate a possible avenue for security. This is due to the fact that the same set of adversarial examples are not misclassified by both classifiers. 

\subsection{Transferability Study Setup}
To properly study the transferability between Vision Transformers, Big Transfer Models and conventional CNNs, we use the same 8 models for CIFAR-10 and CIFAR-100 as mentioned in Section~\ref{subsection:wsetup}. For ImageNet, we also use the same 7 models listed in Section~\ref{subsection:wsetup}. For our transferability study, we consider all possible pairs of classifiers. For each pair of classifiers $(i,j)$, we find a set of $m=1000$ examples that both classifiers correctly identify. We then measure the transferability between the pair of classifiers using Equation~\ref{eq:transfer}. It is important to note that the transferability measurement will be affected by the choice of white-box attack $A_{C_{i}}$ used to generate the adversarial examples. It has been shown that MIM, PGD and FGSM are good candidates for creating highly transferable examples~\cite{mahmood2020beware}. As a result, for every pair of classifiers $(i,j)$, we test all three attacks and report the highest transferability result. For these attacks, we use the same $\epsilon$ and $l_{\infty}$ norm as described in Section~\ref{subsection:wsetup}. Additional experimental details are provided in our supplementary material. 

In Table~\ref{table:transfer}, we show the transferability results for CIFAR-10, CIFAR-100 and ImageNet. The top row of the table corresponds to the model which was used to generate the adversarial examples, $C_{i}$ in Equation~\ref{eq:transfer}. The first column in the table corresponds to the model which was used to predict the labels of the adversarial examples. The model in the first column is $C_{j}$ in Equation~\ref{eq:transfer}. In the special case when $i=j$, we train an independent copy of model $i$ to generate adversarial examples for CIFAR-10 and CIFAR-100. For ImageNet, due to the high computational cost of model training, we forgo the $i=j$ measurement. It can clearly be seen from the other datasets we study and in the literature~\cite{LiuTransfer2017} that copies of the same model $(i=j)$ already have high transferability. We also graphically represent the results of Table~\ref{table:transfer} in Figure~\ref{fig:transferabilityVisual} for the CIFAR-10 dataset. 

\subsection{Analysis of Transferability Study}
\begin{figure}
\includegraphics[scale=0.3]{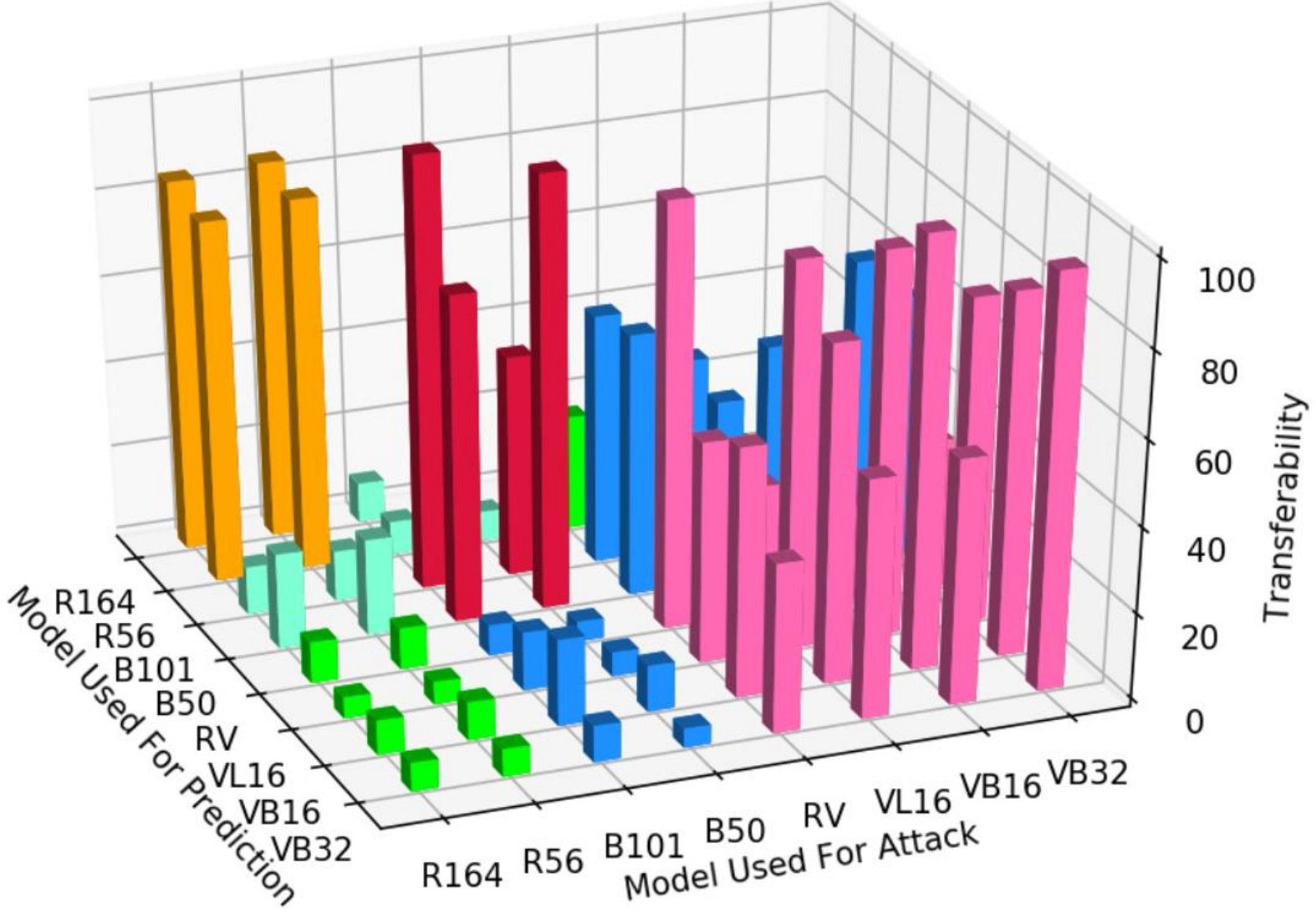}
\caption[]{Visual representation of Table~\ref{table:transfer} for CIFAR-10. The x-axis corresponds to the model used to generate the adversarial examples. The y-axis corresponds to the model used to evaluate the adversarial examples. The z-axis measures transferability between the two models. The bars are color coded based on the two models. Pink, red, and orange bars represent the transferability between models of the same genus. Green, blue and light blue bars represent the transferability measurements between models of different genusus.}
 \label{fig:transferabilityVisual}
 \noindent
\end{figure}

From Table~\ref{table:transfer} and Figure~\ref{fig:transferabilityVisual}, we can see a very interesting phenomenon. The transferability between Vision Transformers and Big Transfer Models is extremely low. For example, consider ViT-L-16 and BiT-M-50x1. Adversarial examples generated using BiT-50x1 are misclassified by ViT-L-16 less than $16\%$ of the time across all datasets ($5.7\%$, $15.5\%$ and $11.8\%$ for CIFAR-10, CIFAR-100 and ImageNet respectively). Likewise, less than half the time BiT-M-50x1 is fooled by adversarial examples generated using ViT-L-16 ($42.5\%$, $47.6\%$ and $34.3\%$ for CIFAR-10, CIFAR-100 and ImageNet). 

Broadly speaking, we can consider the ViT models, BiT models and ResNets each as a model genus. In general, the phenomenon of low transferability mostly occurs between model genusus, but not within model genusus. That is to say, adversarial examples generated by one BiT model will likely transfer to a different BiT model, but not to a ViT model or ResNet. Visually, we can see this result represented for CIFAR-10 in Figure~\ref{fig:transferabilityVisual}. The x-axis represents different models used to generate the adversarial examples and the y-axis represents the model used to evaluate those adversarial examples. The z-axis is used to measure the transferability. For clarity, the bars in the plot are color coded. Green, blue and light blue bars represent the transferability measurements between models of different genusus (green is ViT/ResNet transferability, blue is ViT/BiT transferability and light blue is BiT/ResNet transferability). Pink, red, and orange bars represent the transferability between models of the same genus. Pink is the transferability between ViT models, red is the transferability between BiT models and orange is the transferability between ResNet models.

It is important to note while the low transferability phenomenon is a generally observed trend, it is not an absolute rule. For example, the transferability between Big Transfer models (BiT-M-R50x1 and BiT-M-152x4) for ImageNet is also relatively low ($28\%$ and $24.9\%$). However, the most important factor is that the low transferability phenomenon \textbf{\textit{does}} happen across multiple datasets and for multiple different pairs of models. The usefulness of these observations may not be apparent immediately. Nevertheless, they have serious security implications which we elaborate on subsequently.   
%%%%%%%%%%%%%%%%%%%%%%%Giant table below here look out!

%\caption{Transferability results for CIFAR-10, CIFAR-100 and ImageNet. The first column in each table represents the model used to generate the adversarial examples, $C_{i}$. The top row in each table represents the model used to evaluate the adversarial examples, $C_{j}$. Each entry is the maximum transferability computed using $C_{i}$ and $C_{j}$ over three different attacks, FGSM, PGD and MIM using equation~\ref{eq:transfer}.}

\begin{table*}[]

\caption{Transferability results for CIFAR-10, CIFAR-100 and ImageNet. The first column in each table represents the model used to generate the adversarial examples, $C_{i}$. The top row in each table represents the model used to evaluate the adversarial examples, $C_{j}$. Each entry is the maximum transferability computed using $C_{i}$ and $C_{j}$ over three different attacks, FGSM, PGD and MIM using Equation~\ref{eq:transfer}.}
{\small
\label{table:transfer}
\begin{tabular}{lcccccccc}
 & \multicolumn{8}{c}{CIFAR-10} \\ \cline{2-9} 
\multicolumn{1}{l|}{} & \multicolumn{1}{c|}{ViT-B-32} & \multicolumn{1}{c|}{ViT-B-16} & \multicolumn{1}{c|}{ViT-L-16} & \multicolumn{1}{c|}{R50-ViT} & \multicolumn{1}{c|}{BiT-50x1} & \multicolumn{1}{c|}{BiT-101x3} & \multicolumn{1}{c|}{ResNet-56} & \multicolumn{1}{c|}{ResNet-164} \\ \hline
\multicolumn{1}{|l|}{ViT-B-32} & \multicolumn{1}{c|}{95.8\%} & \multicolumn{1}{c|}{84.1\%} & \multicolumn{1}{c|}{75.5\%} & \multicolumn{1}{c|}{34.9\%} & \multicolumn{1}{c|}{60.8\%} & \multicolumn{1}{c|}{62.0\%} & \multicolumn{1}{c|}{18.6\%} & \multicolumn{1}{c|}{19.9\%} \\ \hline
\multicolumn{1}{|l|}{ViT-B-16} & \multicolumn{1}{c|}{57.1\%} & \multicolumn{1}{c|}{99.6\%} & \multicolumn{1}{c|}{88.9\%} & \multicolumn{1}{c|}{22.6\%} & \multicolumn{1}{c|}{43.4\%} & \multicolumn{1}{c|}{45.0\%} & \multicolumn{1}{c|}{13.9\%} & \multicolumn{1}{c|}{14.0\%} \\ \hline
\multicolumn{1}{|l|}{ViT-L-16} & \multicolumn{1}{c|}{55.6\%} & \multicolumn{1}{c|}{78.4\%} & \multicolumn{1}{c|}{89.6\%} & \multicolumn{1}{c|}{30.3\%} & \multicolumn{1}{c|}{42.5\%} & \multicolumn{1}{c|}{44.7\%} & \multicolumn{1}{c|}{13.0\%} & \multicolumn{1}{c|}{14.8\%} \\ \hline
\multicolumn{1}{|l|}{R50-ViT} & \multicolumn{1}{c|}{39.6\%} & \multicolumn{1}{c|}{58.1\%} & \multicolumn{1}{c|}{51.5\%} & \multicolumn{1}{c|}{98.3\%} & \multicolumn{1}{c|}{61.0\%} & \multicolumn{1}{c|}{58.0\%} & \multicolumn{1}{c|}{26.7\%} & \multicolumn{1}{c|}{29.0\%} \\ \hline
\multicolumn{1}{|l|}{BiT-50x1} & \multicolumn{1}{c|}{4.5\%} & \multicolumn{1}{c|}{10.9\%} & \multicolumn{1}{c|}{5.7\%} & \multicolumn{1}{c|}{4.7\%} & \multicolumn{1}{c|}{100.0\%} & \multicolumn{1}{c|}{51.4\%} & \multicolumn{1}{c|}{7.0\%} & \multicolumn{1}{c|}{9.0\%} \\ \hline
\multicolumn{1}{|l|}{BiT-101x3} & \multicolumn{1}{c|}{8.6\%} & \multicolumn{1}{c|}{20.3\%} & \multicolumn{1}{c|}{13.7\%} & \multicolumn{1}{c|}{7.2\%} & \multicolumn{1}{c|}{75.9\%} & \multicolumn{1}{c|}{100.0\%} & \multicolumn{1}{c|}{7.8\%} & \multicolumn{1}{c|}{9.3\%} \\ \hline
\multicolumn{1}{|l|}{ResNet-56} & \multicolumn{1}{c|}{6.6\%} & \multicolumn{1}{c|}{9.0\%} & \multicolumn{1}{c|}{5.3\%} & \multicolumn{1}{c|}{9.7\%} & \multicolumn{1}{c|}{22.5\%} & \multicolumn{1}{c|}{11.8\%} & \multicolumn{1}{c|}{85.9\%} & \multicolumn{1}{c|}{87.2\%} \\ \hline
\multicolumn{1}{|l|}{ResNet-164} & \multicolumn{1}{c|}{6.8\%} & \multicolumn{1}{c|}{8.1\%} & \multicolumn{1}{c|}{5.0\%} & \multicolumn{1}{c|}{9.7\%} & \multicolumn{1}{c|}{22.3\%} & \multicolumn{1}{c|}{11.2\%} & \multicolumn{1}{c|}{83.6\%} & \multicolumn{1}{c|}{85.7\%} \\ \hline
 & \multicolumn{8}{c}{CIFAR-100} \\ \cline{2-9} 
\multicolumn{1}{l|}{} & \multicolumn{1}{c|}{ViT-B-32} & \multicolumn{1}{c|}{ViT-B-16} & \multicolumn{1}{c|}{ViT-L-16} & \multicolumn{1}{c|}{R50-ViT} & \multicolumn{1}{c|}{BiT-50x1} & \multicolumn{1}{c|}{BiT-101x3} & \multicolumn{1}{c|}{ResNet-56} & \multicolumn{1}{c|}{ResNet-164} \\ \hline
\multicolumn{1}{|l|}{ViT-B-32} & \multicolumn{1}{c|}{96.2\%} & \multicolumn{1}{c|}{88.5\%} & \multicolumn{1}{c|}{83.6\%} & \multicolumn{1}{c|}{52.2\%} & \multicolumn{1}{c|}{60.5\%} & \multicolumn{1}{c|}{61.1\%} & \multicolumn{1}{c|}{14.9\%} & \multicolumn{1}{c|}{14.0\%} \\ \hline
\multicolumn{1}{|l|}{ViT-B-16} & \multicolumn{1}{c|}{71.3\%} & \multicolumn{1}{c|}{99.3\%} & \multicolumn{1}{c|}{93.2\%} & \multicolumn{1}{c|}{38.6\%} & \multicolumn{1}{c|}{44.5\%} & \multicolumn{1}{c|}{47.9\%} & \multicolumn{1}{c|}{9.0\%} & \multicolumn{1}{c|}{7.5\%} \\ \hline
\multicolumn{1}{|l|}{ViT-L-16} & \multicolumn{1}{c|}{67.8\%} & \multicolumn{1}{c|}{88.3\%} & \multicolumn{1}{c|}{94.2\%} & \multicolumn{1}{c|}{48.1\%} & \multicolumn{1}{c|}{47.6\%} & \multicolumn{1}{c|}{50.0\%} & \multicolumn{1}{c|}{9.9\%} & \multicolumn{1}{c|}{9.5\%} \\ \hline
\multicolumn{1}{|l|}{R50-ViT} & \multicolumn{1}{c|}{51.6\%} & \multicolumn{1}{c|}{65.0\%} & \multicolumn{1}{c|}{62.3\%} & \multicolumn{1}{c|}{98.9\%} & \multicolumn{1}{c|}{64.1\%} & \multicolumn{1}{c|}{61.2\%} & \multicolumn{1}{c|}{11.0\%} & \multicolumn{1}{c|}{9.9\%} \\ \hline
\multicolumn{1}{|l|}{BiT-50x1} & \multicolumn{1}{c|}{17.7\%} & \multicolumn{1}{c|}{25.0\%} & \multicolumn{1}{c|}{15.5\%} & \multicolumn{1}{c|}{18.2\%} & \multicolumn{1}{c|}{100.0\%} & \multicolumn{1}{c|}{56.5\%} & \multicolumn{1}{c|}{4.9\%} & \multicolumn{1}{c|}{5.2\%} \\ \hline
\multicolumn{1}{|l|}{BiT-101x3} & \multicolumn{1}{c|}{24.9\%} & \multicolumn{1}{c|}{39.0\%} & \multicolumn{1}{c|}{26.3\%} & \multicolumn{1}{c|}{23.5\%} & \multicolumn{1}{c|}{74.0\%} & \multicolumn{1}{c|}{99.0\%} & \multicolumn{1}{c|}{5.7\%} & \multicolumn{1}{c|}{3.2\%} \\ \hline
\multicolumn{1}{|l|}{ResNet-56} & \multicolumn{1}{c|}{20.1\%} & \multicolumn{1}{c|}{22.2\%} & \multicolumn{1}{c|}{15.3\%} & \multicolumn{1}{c|}{22.7\%} & \multicolumn{1}{c|}{31.4\%} & \multicolumn{1}{c|}{21.9\%} & \multicolumn{1}{c|}{70.8\%} & \multicolumn{1}{c|}{68.9\%} \\ \hline
\multicolumn{1}{|l|}{ResNet-164} & \multicolumn{1}{c|}{22.1\%} & \multicolumn{1}{c|}{24.5\%} & \multicolumn{1}{c|}{15.5\%} & \multicolumn{1}{c|}{24.2\%} & \multicolumn{1}{c|}{35.9\%} & \multicolumn{1}{c|}{26.5\%} & \multicolumn{1}{c|}{74.5\%} & \multicolumn{1}{c|}{79.2\%} \\ \hline
 & \multicolumn{8}{c}{ImageNet} \\ \cline{2-8}
\multicolumn{1}{l|}{} & \multicolumn{1}{c|}{ViT-B-16} & \multicolumn{1}{c|}{ViT-L-16} & \multicolumn{1}{c|}{ViT-L-16 (512)} & \multicolumn{1}{c|}{BiT-50x1} & \multicolumn{1}{c|}{BiT-152x4} & \multicolumn{1}{c|}{ResNet-50} & \multicolumn{1}{c|}{ResNet-152} & \multicolumn{1}{l}{} \\ \cline{1-8}
\multicolumn{1}{|l|}{ViT-B-16} & \multicolumn{1}{c|}{+} & \multicolumn{1}{c|}{89.1\%} & \multicolumn{1}{c|}{39.6\%} & \multicolumn{1}{c|}{40.8\%} & \multicolumn{1}{c|}{27.4\%} & \multicolumn{1}{c|}{44.0\%} & \multicolumn{1}{c|}{40.1\%} & \multicolumn{1}{l}{} \\ \cline{1-8}
\multicolumn{1}{|l|}{ViT-L-16} & \multicolumn{1}{c|}{90.9\%} & \multicolumn{1}{c|}{+} & \multicolumn{1}{c|}{64.5\%} & \multicolumn{1}{c|}{40.0\%} & \multicolumn{1}{c|}{26.9\%} & \multicolumn{1}{c|}{43.7\%} & \multicolumn{1}{c|}{40.8\%} & \multicolumn{1}{l}{} \\ \cline{1-8}
\multicolumn{1}{|l|}{ViT-L-16 (512)} & \multicolumn{1}{c|}{28.0\%} & \multicolumn{1}{c|}{43.4\%} & \multicolumn{1}{c|}{+} & \multicolumn{1}{c|}{34.3\%} & \multicolumn{1}{c|}{26.3\%} & \multicolumn{1}{c|}{28.4\%} & \multicolumn{1}{c|}{23.2\%} & \multicolumn{1}{l}{} \\ \cline{1-8}
\multicolumn{1}{|l|}{BiT-50x1} & \multicolumn{1}{c|}{9.8\%} & \multicolumn{1}{c|}{8.4\%} & \multicolumn{1}{c|}{11.8\%} & \multicolumn{1}{c|}{+} & \multicolumn{1}{c|}{24.9\%} & \multicolumn{1}{c|}{24.7\%} & \multicolumn{1}{c|}{18.7\%} & \multicolumn{1}{l}{} \\ \cline{1-8}
\multicolumn{1}{|l|}{BiT-152x4} & \multicolumn{1}{c|}{8.2\%} & \multicolumn{1}{c|}{7.6\%} & \multicolumn{1}{c|}{13.5\%} & \multicolumn{1}{c|}{28.0\%} & \multicolumn{1}{c|}{+} & \multicolumn{1}{c|}{15.1\%} & \multicolumn{1}{c|}{12.0\%} & \multicolumn{1}{l}{} \\ \cline{1-8}
\multicolumn{1}{|l|}{ResNet-50} & \multicolumn{1}{c|}{23.8\%} & \multicolumn{1}{c|}{18.8\%} & \multicolumn{1}{c|}{24.7\%} & \multicolumn{1}{c|}{55.3\%} & \multicolumn{1}{c|}{24.4\%} & \multicolumn{1}{c|}{+} & \multicolumn{1}{c|}{86.7\%} & \multicolumn{1}{l}{} \\ \cline{1-8}
\multicolumn{1}{|l|}{ResNet-152} & \multicolumn{1}{c|}{25.9\%} & \multicolumn{1}{c|}{22.1\%} & \multicolumn{1}{c|}{26.6\%} & \multicolumn{1}{c|}{54.1\%} & \multicolumn{1}{c|}{26.8\%} & \multicolumn{1}{c|}{89.4\%} & \multicolumn{1}{c|}{+} & \multicolumn{1}{l}{} \\ \cline{1-8}
\end{tabular}
}
\end{table*}

\section{White-Box Security and Transferability}
\label{sec:ensemble}
\textit{How can the transferability phenomena be leveraged to provide security?} From section~\ref{sec:transfer}, we know that the transferability of adversarial examples between different model genusus is generally low. Therefore, we propose testing an ensemble of different models as a defense. To further clarify the original question, we break it down into two parts: Can an ensemble defense provide security against a white-box adversary, and can an ensemble provide security against a black-box adversary? In this section we answer the white-box question by proposing a novel attack that simultaneously breaks both Transformers and CNNs. In Section~\ref{sec:bb}, we investigate the black-box question. 

We first define our base case ensemble defense.
    %\item \textit{} In section~\ref{sec:bb} we show that under black-box adversaries, an ensemble of Vision Transformers and Big Transfer Models provide security without any drop in clean accuracy. This important discovery highlights exactly how Vision Transformers \textit{advance} the field of adversarial machine learning.
%\end{enumerate}

\textbf{Ensemble Models:} In this paper, we have already examined multiple Vision Transformers, Big Transfer Models and ResNets. The simplest ensemble would be to choose two types of classifiers from this group. Therefore, as a base case we use the most complex BiT model and ViT model. For CIFAR-10 and CIFAR-100 datasets, the ensemble is comprised of ViT-L-16 and BiT-M-101x3. For ImageNet, this ensemble is made up of ViT-L-16 (image size 512) and BiT-M-152x4. Here, we do not consider ResNets as they have significantly less clean accuracy and we do not want to pay such a security cost. In the supplementary material, we do provide some ResNet ensemble experiments for the sake of completeness. 

\textbf{Ensemble Output:} In our ensemble defense, there are several possible ways to combine the output of the models. Here we consider three ways commonly found in the literature, majority voting~\cite{pang2019improving}, absolute consensus~\cite{mahmood2020buzz} and random selection~\cite{srisakaokul2018muldef}. 
Majority voting is a weak method of evaluating adversarial examples because not every classifier must be fooled, resulting in diminishing returns as the number of classifiers increases. The alternative to majority voting is absolute consensus~\cite{mahmood2020buzz}. In this setup, if every classifier does not agree on the same class label then the sample is marked as adversarial. Absolute consensus removes the diminishing returns disadvantage of majority voting, though at the cost of clean accuracy. In absolute consensus, it is typical that many clean samples are marked as adversarial~\cite{mahmood2020buzz}. Due to this, we use random selection in all our ensemble defenses for the remainder of the paper. In random selection, a single model is selected randomly and used to evaluate the input at run time.  

% To illustrate the diminishing returns of majority voting, consider a case of $100$ classifiers: only $51$ must be fooled for the attack to succeed in a majority voting based technique.

\subsection{The Self-Attention Gradient Attack}
\label{subsec:saga}
\textbf{Attack Motivation:} A naïve approach would be to assume that an ensemble defense would provide security against a white-box adversary if only the low transferability results in Section~\ref{sec:whitebox} and Section~\ref{sec:transfer} were taken into account. Consider the following analysis: Let us focus on the ImageNet models ViT-L-16 (image size 512) and BiT-M-152x4. From Section~\ref{sec:transfer}, we know a white-box MIM attack has a $100\%$ attack success rate ($0\%$ robust accuracy) on ViT-L-16 (see Table~\ref{table:transfer}). Now let us introduce an additional model, BiT-M-152x4 into the ensemble with ViT-L-16. From Section~\ref{sec:transfer} Table~\ref{table:transfer}, we know the adversarial examples generated from ViT-L-16 will be misclassified by BiT-M-152x4 only $26.3\%$ of the time. If we make an ensemble of ViT-L-16 and BiT-M-152x4 with random selection, this means the attack success rate on average would drop to $63.15\%$. It seems as if we went from $0\%$ robust accuracy using only ViT-L-16 to $36.85\%$ robust accuracy just by using an ensemble with random selection. However, this is not the case as the adversarial examples we are using only come from attacking one model. We demonstrate the flaws in this type of analysis by proposing a new attack which generates adversarial examples that are simultaneously misclassifed by both Vision Transformers and CNNs. We call this new attack, the Self-Attention Gradient Attack (SAGA).

\textbf{Mathematical Description:} To derive SAGA, we assume the same white-box adversary we detailed in Section~\ref{sec:whitebox}. Such an adversary has knowledge of the models and trained parameters in an ensemble defense. Instead of focusing completely on optimizing over one of the models, SAGA focuses on breaking multiple models at once. Assume we are given an ensemble with a set of Vision Transformers $V$ and a set of CNNs $K$. The goal of the attacker is to craft an adversarial example $x_{adv}$ from $x$ within perturbation bounds $\epsilon$ that is misclassifed by all members $v \in V$ and $k \in K$. We can iteratively compute the adversarial example as follows:
\begin{equation}
    x_{adv}^{(i+1)} = x_{adv}^{(i)} + \epsilon_{s}*sign(%\nabla
    G_{blend}(x_{adv}^{(i)}))
\end{equation}
where $x_{adv}^{(1)}=x$ and $\epsilon_{s}$ is the step size for the attack. Further, we define $%\nabla
G_{blend}(x_{adv}^{(i)})$ as follows:
\begin{equation}
\label{eq:blend}
    G_{blend}(x_{adv}^{(i)}) = \sum_{k \in K} \alpha_{k} \frac{\partial L_{k}}{\partial x_{adv}^{(i)}} + \sum_{v \in V} \alpha_{v} \phi_{v} \odot \frac{\partial L_{v}}{\partial x_{adv}^{(i)}}
\end{equation}
In Equation~\ref{eq:blend}, the first summation is for the models in set $K$ which are CNNs. $\partial L_{k}/\partial x_{adv}^{(i)}$ is the partial derivative of the loss function of the $k^{th}$ CNN with respect to the adversarial input $x_{adv}^{(i)}$. Each model $k$ has an associated weighting factor $\alpha_{k}$. In a more refined approach, $\alpha_{k}$ could be optimized over as well, but here we simply leave $\alpha_{k}$ as a hyperparameter in the attack. Note that PGD~\cite{PGDmadry2018} without randomized start is a special case of our attack when $V=\emptyset$, $K$ has exactly one element and $a_{1}=1$. However, when attacking an ensemble, $V\neq \emptyset$ and hence we have a second term.

%$V \in \{\emptyset\}$ [TO DO you mean $V=\emptyset$],  However, when attacking an ensemble $V \notin \{\emptyset\}$ [TO DO You mean $V\neq \emptyset$] and hence we have a second term.

In Equation~\ref{eq:blend}, the second term\footnote{$\odot$ is the element wise Hadamard product; $x$ in (\ref{eq:blend}) and (\ref{eq:phiv}) is an image matrix and the partial derivative w.r.t $x$ in (\ref{eq:blend}) is represented as a matix.} $\alpha_{v} \phi_{v} \odot \partial L_{v}/\partial x_{adv}^{(i)}$ is used to craft adversarial examples that are misclassified by the Vision Transformers in the ensemble. Here $\partial L_{v}/\partial x_{adv}^{(i)}$ is the loss function of the transformer with respect to the adversarial input. Likewise, $\alpha_{v}$ is a weighting factor selected by the attacker to balance the emphasis on different models. We also bring in one additional term which is specific to Vision Transformers, $\phi_{v}$. The term $\phi_{v}$ is the self-attention map associated with the $v^{th}$ transformer in the ensemble. 

The self-attention $\phi_{v}$ is computed using attention rollout~\cite{abnar2020quantifying} and is defined as:
\begin{equation}
    \phi_{v}= \left(
    \prod_{l=1}^{n_l} \left[ \sum_{i=1}^{n_h} (0.5W_{l,i}^{(att)}+0.5 I) \right] \right) \odot x. \label{eq:phiv}
\end{equation}
where $n_{h}$ is the number of attention heads per layer, $n_{l}$ is the number of attention layers, $W_{l,i}^{(att)}$ is the attention weight matrix in each attention head, $I$ is the identity matrix and $x$ is the input image. This technique takes into account the attention flow from each layer of the transformer to the next layer, including the effect of skip connections. The attention values from the different attention heads within the same layer are averaged, and the attention values are recursively multiplied between different layers.

\begin{figure}
\includegraphics[scale=0.7]{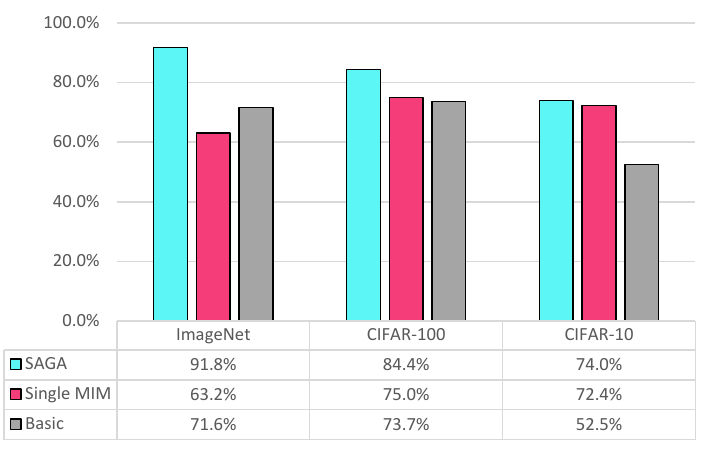}
\caption[]{Attack success rate of the Self-Attention Gradient Attack (SAGA), the Single MIM attack and Basic attack on an ensemble containing one ViT-L-16 model and one BiT-M-R101x3 model (or BiT-M-R152x4 for ImageNet). For full descriptions of each attack see Section~\ref{subsec:saga}.}
 \label{fig:SAGA}
 \noindent
\end{figure}

\textbf{Experimental Results:} We demonstrate the SAGA results by attacking a simple ensemble of Vision Transformers and Big Transfer Models for CIFAR-10, CIFAR-100 and ImageNet. We use $1000$ clean correctly identified examples with the same attack parameters as described in Section~\ref{sec:whitebox}. For CIFAR-10 and CIFAR-100, we use Bit-M-R101x3 and ViT-L-16. For ImageNet, we use Bit-M-R152x4 and ViT-L-16. We also test two other simple attacks which are denoted in Figure~\ref{fig:SAGA} as 'Basic' and 'Single MIM'. The basic attack is a combination of the model gradients without weighted coefficients and self-attention included. The single MIM attack is the best transfer attack on the ensemble as reported from Table~\ref{table:transfer}. Here we use MIM as opposed to FGSM or PGD in the 'Single MIM' attack as we experimentally found MIM samples to transfer better. 

The main contribution of this attack is to demonstrate that Vision Transformer/Big Transfer type of ensembles are not secure under a white-box adversary. This is precisely what is shown in Figure~\ref{fig:SAGA}. SAGA has an attack success rate of $74.0\%$, $84.4\%$ and $91.8\%$ on the ensemble for CIFAR-10, CIFAR-100 and ImageNet, respectively. In Figure~\ref{fig:SAGA}, we also show SAGA outperforms the two other white-box multi-model attacks across all datasets. For brevity, many details are omitted here such as the hyperparameter selection for SAGA and attacks on Transformer/ResNet ensembles. We provide this information fully in the supplementary material.

%As these matters are not as crucial to our main work, we relegate them to the supplementary material for those interested.

%The most important result is shown here, that a transformer/CNN white-box ensemble is not secure. However, 
\section{Black-Box Security and Transferability}
\label{sec:bb}
In this section, we consider the transferability phenomena and its security implications under a black-box adversarial model. We once again use an ensemble of classifiers with random selection as described in Section~\ref{sec:ensemble}. From Section~\ref{subsec:saga}, we know that such an ensemble is not secure against white-box adversaries. Using attacks like SAGA, an adversary can blend the gradients of different models and the self-attention of Transformers. This results in a high percentage of adversarial examples that are misclassified by all the classifiers. However, this type of attack relies heavily on the white-box capabilities of the adversary. Without knowledge of the models in the ensemble and their trained parameters, this type of attack would not work. This brings up a new possibility. Can transferability (through an ensemble) provide security when individual model gradients are not available to the attacker? 
\subsection{Black-Box Attack Parameters and Adversarial Model}
 \textbf{Adversarial Model:} In this section we consider two of the main types of black-box adversaries, query-based~\cite{brendel2018decisionbased} and transfer-based adversaries~\cite{papernot2017practical}. For the query based adversary, we test one of the most recent attacks, the RayS attack~\cite{chen2020rays}. In this attack, the adversary generates an adversarial example by repeatedly querying the defense and adjusting the noise accordingly. For the transfer attack, we implement the Adaptive Black-Box Attack~\cite{mahmood2020buzz}. This attack is a stronger version of the Papernot attack originally proposed in~\cite{papernot2017practical}. Here the attacker has access to a percentage of the original training data, query access to the defense and the ability to train a synthetic model to generate adversarial examples. In this attack, the adversary queries the defense to obtain labels for the training data. It then uses the data labeled by the defense to train an independent classifier (synthetic model). A white-box attack is then performed on the trained synthetic model. The resulting adversarial examples are then tested on the defense. 

\textbf{Attack Parameters:} For all black-box attacks, we use the same basic set of constraints as described in section~\ref{subsection:wsetup}. The noise the adversary can generate is bounded by the $l_{\infty}$ norm with $\epsilon=0.031$ for CIFAR-10/CIFAR-100 and $\epsilon=0.062$ for ImageNet. For the RayS attack, we give the adversary a budget of $10,000$ queries per sample. For the Adaptive attack, we give the adversary $100\%$ of the training data. For the synthetic model in this attack, we used ViT-B-32 pre-trained on ImageNet-21K. We also experimented with CNN based synthetic models, however these did not perform as well on our ensemble defense. It should also be noted the $100\%$ strength attack requires a huge amount of computation. Due to this we only show the results for CIFAR-10 for the Adaptive attack. For RayS, we show results for all three datasets.

\begin{figure}
\includegraphics[scale=0.6]{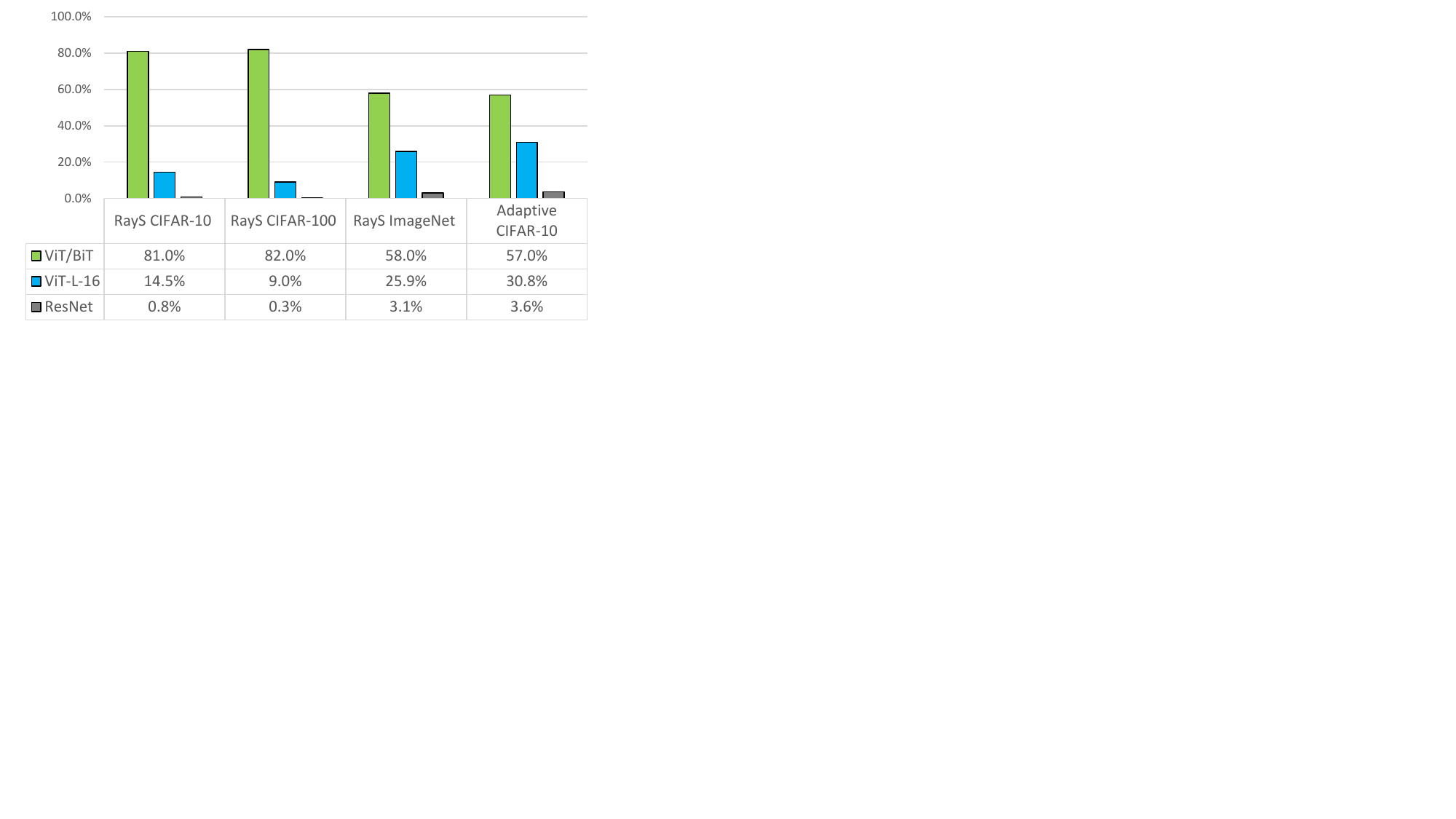}
\caption[]{Robust accuracy (higher is better) of different model configurations under black-box attacks. Here ViT/BiT is an ensemble containing a Vision Transformer (ViT-L-16) and a Big Transfer Model (BiT-M-101x3 for CIFAR-10/CIFAR-100 and Bit-M-R152x4 for ImageNet.}
 \label{fig:bb}
 \noindent
\end{figure}

\subsection{Black-Box Attack Analysis}
In Figure~\ref{fig:bb}, we show the results graphically for the RayS and Adaptive attack. We consider three different model configurations. We test an ensemble of one Vision Transformer (ViT-L-16) and one Big Transfer Model (BiT-M-101x3 for CIFAR-10/CIFAR-100 and BiT-M-152x4 for ImageNet). We also test a single ViT-L model and a single CNN (ResNet-56 for CIFAR-10/CIFAR-100 and ResNet-50 for ImageNet). While slightly redundant, we do test other ensemble configurations (and individual Big Transfer Models) in the supplementary material for those interested. 

The robust accuracy (percent of adversarial samples correctly identified by the defense) are shown in Figure~\ref{fig:bb} for each attack. Here we observe the most significant result of our paper: a simple ensemble including a Vision Transformer and Big Transfer model drastically improves security. For RayS, we observe an increase of $66.5\%$, $73\%$ and $32.1\%$ in robust accuracy for CIFAR-10, CIFAR-100 and ImageNet respectively. For the CIFAR-10 Adaptive attack, even when the adversary has $100\%$ of the training data, query access and a synthetic model pre-trained on the same dataset as the defense (ImageNet-21K), we can still achieve a robust accuracy of $57\%$. For the Adaptive attack that represents an improvement of $26.2\%$ over a single model. 

We also stress that this improvement does not come at the cost of clean accuracy. The average clean accuracy of the ensemble is $98.2\%$, $92.83\%$ and $85.37\%$ for CIFAR-10, CIFAR-100 and ImageNet respectively. By leveraging the low transferability phenomena we previously studied, we are able to create a defense that achieves near state-of-the-art performance on clean data and gives significant black-box robustness. 

\section{Conclusion}
\label{sec:conclusion}

The introduction of Vision Transformers represents new opportunities for the field of adversarial machine learning. By analyzing these new models, we are the first to uncover several intriguing properties: First, using six different attacks, we showed individual Vision Transformers are just as vulnerable as their CNN counterparts to white-box adversaries. Second, we studied the transferability of adversarial examples between Vision Transformers and other model genusus using eight different models for CIFAR-10/CIFAR-100 and seven different models for ImageNet. 

We demonstrated that the transferability between different model genusus are in general, remarkably low. We then showed this phenomena does not yield white-box security by developing a new white-box attack, the Self-Attention Gradient Attack (SAGA). Finally, we showed that under a black-box adversary, the transferability phenomena can be used to achieve robustness. Using a two model ensemble, we demonstrated robustness to black-box adversarial attacks. This included improving the robust accuracy by as much as $60\%$ in some cases, all while maintaining near state-of-the-art clean accuracy on CIFAR-10, CIFAR-100 and ImageNet. Through our comprehensive experiments and analyses, we show how Vision Transformers advance security in the field of adversarial machine learning.

%\clearpage

{\small
\bibliographystyle{ieee_fullname}
\bibliography{OurRef}

%\bibliography{egbib}
}
\clearpage
\begin{center}
\textsc{\Large On the Robustness of Vision Transformers to Adversarial Examples Supplemental Material}
\end{center}
\appendix
\section{Supplementary Material Organization}

In this section, we briefly describe the organization of our supplementary material so that readers can find  pertinent material with ease. Overall our supplementary material provides more white-box experiments, more black-box experiments, results for adversarial training of Vision Transformers and further investigation of the transferability phenomena observed in the main paper.

\textbf{White-Box and Black-Box Attacks:} We start with full descriptions of the white-box attack used in this paper and their corresponding parameters in section~\ref{sup:whitebox}. In this section, we also provide the CIFAR-100 white-box attack results not given in the main paper (due to brevity as well as redundancy). Aside from the conventional white-box attacks, we also give more details for the Self-Attention Gradient Attack (SAGA) in section~\ref{sup:saga}. We follow up our white-box type of attack sections with a full description of the black-box adversarial model and attack parameters in section~\ref{sup:blackbox}. In this section we also include black-box attack experimental results on individual models (Vision Transformers, Big Transfer Models and ResNets) for RayS. In addition, we further provide a series of hyperparameter experiments used to fine-tune the transfer based black-box attack (also known as the adaptive black-box attack). The results of these hyperparameter experiments reveal some very interesting implications for the design of future black-box attacks. 

\textbf{Decision Region Graphs and Transferability:} In section~\ref{sup:whitebox_transformer} we delve further into the transferability phenomena. We start by discussing recent work that mathematical shows the equivalence between Transformers and CNNs. We then empirical show how the conditions under which this equivalence happens does not likely occur for Vision Transformers and other CNNs. We demonstrate this empirical by graphing the decision regions for different Vision Transformers and CNNs for CIFAR-10, CIFAR-100 and ImageNet.

\textbf{Adversarial Trained Vision Transformers:} One topic that is a major part of the field of adversarial machine learning, but we didn't have space to cover is adversarial training. Can Vision Transformers be adversarial trained as well as CNNs? We answer this question by experimenting with different adversarial training techniques on Vision Transformers and CNNs for CIFAR-10 and CIFAR-100 in section~\ref{sup:FAT_defense}. Lastly in section~\ref{sup:tables} we provide some additional numerical tables for our results that are not of extreme importance, but may assist anyone wishing to replicate our results.

\section{White-Box Attacks}
\label{sup:whitebox}

In this section, we mathematically define the white-box adversarial model. We also give detailed descriptions of the white-box attacks tested in this paper. Along with the attack descriptions, we also list the parameters we choose for each white-box attack. 

\subsection{White-Box Adversarial Model}

\textbf{Mathematical Description:} Formally, we can mathematically describe our adversarial model follows: We start with a  classifier $C$ with (trained) parameters $\theta$. Given input $x$, the classifier outputs label $y$ such that $C(x,\theta) = y$. The goal of the adversary is to create an adversarial sample $x_{adv}$ from $x$ such that:

\begin{equation}
\label{eq:condition}
C(x_{adv}, \theta) \neq y 
\end{equation}
where $x_{adv}$ is created from $x$ using attack $A$. That is $x_{adv} \triangleq A(x)$ and $x_{adv}$ is subject to the following constraint:
\begin{equation}
\label{eq:condition2}
\|x-x_{adv}\|_{p} \leq \epsilon
\end{equation}
where $p$ is the type of norm used to measure the distance between $x$ and $x_{adv}$ and $\epsilon$ is the maximum allowed distance between $x$ and $x_{adv}$. Finally, there is one additional constraint. The image must be within a valid pixel range:

\begin{equation}
\label{eq:condition3}
    x \in [p_{min},p_{max}]^{n \times m \times r}
\end{equation}
where in equation~\ref{eq:condition3}, $p_{min}$ and $p_{max}$ refer to the minimum and maximum pixel values of a valid image, $n$ and $m$ refer to the size of the image and $r$ represents the number of color channels in the image.

\textbf{Adversarial Capabilities:} In the white-box adversarial model the adversary has knowledge of $C$, $\theta$, $x$ and $y$. Here $C$ represents the type of classifier (e.g. CNN) and classifier architecture (e.g. ResNet-56). They also know the trained parameters of the classifier $\theta$. For a CNN or Transformer this would be the weights and biases of the classifier model. Lastly, the adversary has a clean example $x$ and corresponding class label $y$. In this paper, we focus on the untargeted attack model. This means the adversary succeeds if and only if equation~\ref{eq:condition}, equation~\ref{eq:condition2} and equation~\ref{eq:condition3} all hold true. 

\subsection{Types of White-Box Attacks}

In a white-box attack, the adversary crafts $x_{adv}$ from $x$ using technique $A$. The choice of $A$ can heavily affect the success rate of the attack (the percent of samples that are misclassified by $C$). From the literature, there are many different techniques to craft white-box attacks. Below we describe each of the attacks we test in this paper: 

\begin{enumerate}
    \item \textbf{Fast Gradient Sign Method} - The Fast Gradient Sign Method (FGSM)~\cite{FGSMpaper} creates adversarial examples through the addition of non-random noise in the direction of the gradients of the loss function:
    \begin{equation}
    \label{eq:fgsm}
    x_{adv} = x + \epsilon*sign(\nabla_x L(x,y;\theta))
    \end{equation}
    where \emph{L} is the loss function of the classifier $C$. Note that in Equation~\ref{eq:fgsm}, only the sign of the loss function is used and the magnitude of the second term is dictated by $\epsilon$, which is a small perturbation added to the image $x$. It is also important to note that this is a single step attack. The adversary only backpropagates on the model once to obtain the gradient of the loss function and then applies this directly to $x$.
    
    \item \textbf{Projected Gradient Descent} - Projected Gradient Descent (PGD)~\cite{PGDmadry2018} is a multi-step variant of the FGSM algorithm. It attempts to find the minimum bounded perturbation that maximizes the loss of a model through initializing a random perturbation in a ball of radius \emph{d} with center \emph{x}. A gradient step is taken in the direction of the greatest loss and the perturbation is then projected back into this ball. The $k$-step PGD algorithm initializes $x^0 = x$ and the perturbed image $x^i$ in the $i^{th}$ step is computed as:
    \begin{equation}
    \label{eq:pgd}
    x^i = P(x^{i-1} + \alpha*sign(\nabla_x L(x^{i-1};y;\theta)))
    \end{equation}
    where $P$ 
is the projection function that projects the adversarial data back into the $\epsilon$-ball centered at $x^{i-1}$ if necessary, and $\alpha$ is the step size. The bounds on the projection are defined by the $l_{p}$ norm.% or $l_{2}$. For example, in a PGD-20 attack, 20 iterations of Equation~\ref{eq:pgd} are used and the maximum perturbation in a step $\alpha$ can be $0.031/20$ for CIFAR-10.
    
    \item \textbf{Backward Pass Differentiable Approximation} - Backward Pass Differentiable Approximation (BPDA)~\cite{BPDApaper} is an attack designed to overcome non-differentiable functions that would ordinarily prevent the use of backpropagation to generate adversarial examples. BPDA is capable of creating effective adversarial examples for those cases in which the defense employs gradient masking or another technique in which the gradient is obfuscated. The gradient can be obfuscated in one of three ways: shattered gradients, stochastic gradients, and exploding/vanishing gradients. Shattered gradients in a defense either introduce numerical instability or cause a gradient to be nonexistent or incorrect~\cite{BPDApaper}. Stochastic gradients are generally a result of randomized defenses. Exploding and vanishing gradients generally occur in recurrent neural networks.
    
    For a neural network $f(\cdot)=f^{1...j}$ with a non-differentiable layer $f^i(\cdot)$, the first step of BPDA is to find a differentiable function \emph{g(x)} that approximates $f^i$. The gradient of the network \emph{f}, $\nabla_x f(x)$, is then approximated by performing a forward pass through $f(\cdot)$, and then only on the backward pass replacing $f^i(x)$ by $g(x)$.
    Adversarial examples are generated using a similar approach to PGD~\cite{PGDmadry2018}.
    
    \item \textbf{Momentum Iterative Method} - A subset of gradient descent approaches, the Momentum Iterative Method (MIM)~\cite{MIMdong2018boosting} applies a velocity vector in the direction of the gradient of the loss function across iterations. 
    Momentum is used to create the gradient:
     \begin{equation}
    \label{momentum}
    x'_{i} = clip_{x,\epsilon}(x'_{i-1} + \frac{\epsilon}{r}*sign(g(i)))  
    \end{equation}
    Since MIM takes into account previous gradients, it is able to overcome narrow valleys, small bumps, and local minima and maxima. Specifically, the momentum algorithm gathers the gradients of $t$ iterations with a decay factor $\mu$. The adversarial example $x^*_{t}$ is perturbed in the direction of the accumulated gradient with a step size of $\alpha$. Note that if $\mu=0$, the MIM algorithm degenerates to iterative FGSM.
    \\For targeted attacks, the aim of finding an adversarial example misclassified as a target class $y^*$ is to minimize the loss function $J(x^*,y^*)$. The accumulated gradient is derived as follows:
    \begin{equation}
    \label{eq:mim}
    g_{t+1} = \mu * g_t + \frac{J(x^*_{t},y^*)}{||\nabla_x J(x^*_{t},y^*)||_{1}}
    \end{equation}
    For a targeted attack with an $L_{\infty}$ norm bound, the adversarial computation becomes:
    \begin{equation}
    \label{eq:mim_fgsm}
    x^*_{t+1} = x^*_{t} - \alpha * sign(g_{t+1})
    \end{equation}
    
    \item \textbf{Carlini and Wagner Attack} - The aim of the Carlini and Wagner ($C\&W$) attack ~\cite{CWAttackCarliniWagner2016} is to perturb an image by a minimal delta such that the image will be misclassified. The following objective function is used to find the adversarial noise:
     \begin{align}
    \label{eq:cw}
    %&min_{\omega}||x'(\omega)-x||_{2}^2 + cf(x'(\omega)) \\
    &\min||\frac{1}{2}(\tanh(\omega)+1) - x ||_{2}^2 + c\cdot f (\frac{1}{2}(\tanh(\omega)+1))\\
    \label{eq:cw2}
    &f(x') = \max(\max \{Z(x')_{i}: i \neq t\} - Z(x')_{t},-\kappa)
    \end{align}
    where \emph{f} is the best objective function, $\omega$ is the perturbation, $t$ is the chosen target class, $\kappa$ is a constant that controls the confidence with which the sample is misclassified, $Z(x')$ is the output from the logits layer, and \emph{c} is a constant chosen through binary search.
    $C\&W$ is an iterative attack because the objective of the $C\&W$ attack is formulated as an optimization problem, as given by Equation~\ref{eq:cw}. 
    %It requires many iterations to obtain the adversarial example.
    
    \item \textbf{Auto Projected Gradient Descent} - Auto Projected Gradient Descent (APGD)~\cite{croce2020reliable} is an automated version of PGD in which the step size is not fixed, but instead changes adaptatively. In APGD, the total iterations are divided into an exploration phase and an exploitation phase. A larger step size is used in the former phase, allowing for quicker exploration, while a smaller step size is used in the latter phase to fine-tune the maximization of the loss function. The choice of step size in APGD is determined by a budget of $N_{iter}$ iterations and the cumulative progress of optimization, as defined by two conditions in equations~\ref{eq:apgd} and ~\ref{eq:apgd2}:
   \begin{flalign} 
   \label{eq:apgd}
    &\sum_{i=w_{j-1}}^{w_j -1} \textbf{1}_{f(x^{(i+1)})>f(x^{(i)})}<\rho*(w_j - w_{j-1}), &\\
     \label{eq:apgd2}
    &\eta^{(w_{j -1})} \equiv  \eta^{(w_{j})} \land f_{max}^{(w_{j -1})} \equiv f_{max}^{(w_{j})}&
    \end{flalign}
    where $w_j$ are the checkpoints at which the algorithm can reduce the step size by a factor of 2 and $ f_{max}^k$ is the highest objective value reached in the first $k$ iterations. If one of the above two conditions is met, then the step size at iteration $k=w_j$ is halved and $\eta^{(k)} := \eta^{(w_{j})} / 2$ for every $k = w_j + 1, ... , w_{j+1}$. A version of the Auto-PGD which uses cross-entropy is referred to as APGD-CE. This attack was shown to be the best performing attack of the different APGD variatons~\cite{croce2020reliable}. Therefore, we use APGD-CE in our white-box attacks.

\end{enumerate}

\begin{table*}[]
\caption{White-box attacks on Vision Transformers, Big Transfer Models and ResNets. The attacks are done using the $l_{\infty}$ norm with $\epsilon=0.031$ for CIFAR-10 and $\epsilon=0.062$ for ImageNet. In this table the robust accuracy is given for each corresponding attack. The last column "Acc" refers to the clean accuracy of the model. In the main paper part of this table was also presented (see table~\ref{table:whiteboxattacks}) but without CIFAR-100 results for brevity. The table here represents the full white-box attack results.}
\centering
{\small
\begin{tabular}{lccccccc}
 & \multicolumn{7}{c}{CIFAR-10} \\ \cline{2-8} 
\multicolumn{1}{l|}{} & \multicolumn{1}{c|}{FGSM} & \multicolumn{1}{c|}{PGD} & \multicolumn{1}{c|}{BPDA} & \multicolumn{1}{c|}{MIM} & \multicolumn{1}{c|}{C\&W} & \multicolumn{1}{c|}{APGD} & \multicolumn{1}{c|}{Acc} \\ \hline
\multicolumn{1}{|l|}{ViT-B-32} & \multicolumn{1}{c|}{37.9\%} & \multicolumn{1}{c|}{1.8\%} & \multicolumn{1}{c|}{17.6\%} & \multicolumn{1}{c|}{4.4\%} & \multicolumn{1}{c|}{0.0\%} & \multicolumn{1}{c|}{0.0\%} & \multicolumn{1}{c|}{98.6\%} \\ \hline
\multicolumn{1}{|l|}{ViT-B-16} & \multicolumn{1}{c|}{39.5\%} & \multicolumn{1}{c|}{0.0\%} & \multicolumn{1}{c|}{20.3\%} & \multicolumn{1}{c|}{0.3\%} & \multicolumn{1}{c|}{0.0\%} & \multicolumn{1}{c|}{0.0\%} & \multicolumn{1}{c|}{98.9\%} \\ \hline
\multicolumn{1}{|l|}{ViT-L-16} & \multicolumn{1}{c|}{56.3\%} & \multicolumn{1}{c|}{1.2\%} & \multicolumn{1}{c|}{28.7\%} & \multicolumn{1}{c|}{5.9\%} & \multicolumn{1}{c|}{0.0\%} & \multicolumn{1}{c|}{0.0\%} & \multicolumn{1}{c|}{99.1\%} \\ \hline
\multicolumn{1}{|l|}{ViT-R50} & \multicolumn{1}{c|}{40.8\%} & \multicolumn{1}{c|}{0.1\%} & \multicolumn{1}{c|}{13.4\%} & \multicolumn{1}{c|}{0.2\%} & \multicolumn{1}{c|}{0.0\%} & \multicolumn{1}{c|}{0.0\%} & \multicolumn{1}{c|}{98.6\%} \\ \hline
\multicolumn{1}{|l|}{BiT-M-R50x1} & \multicolumn{1}{c|}{66.0\%} & \multicolumn{1}{c|}{0.0\%} & \multicolumn{1}{c|}{14.9\%} & \multicolumn{1}{c|}{0.0\%} & \multicolumn{1}{c|}{0.0\%} & \multicolumn{1}{c|}{0.0\%} & \multicolumn{1}{c|}{97.5\%} \\ \hline
\multicolumn{1}{|l|}{BiT-M-R101x3} & \multicolumn{1}{c|}{85.2\%} & \multicolumn{1}{c|}{0.0\%} & \multicolumn{1}{c|}{17.1\%} & \multicolumn{1}{c|}{0.0\%} & \multicolumn{1}{c|}{0.0\%} & \multicolumn{1}{c|}{0.0\%} & \multicolumn{1}{c|}{98.7\%} \\ \hline
\multicolumn{1}{|l|}{ResNet-56} & \multicolumn{1}{c|}{23.0\%} & \multicolumn{1}{c|}{0.0\%} & \multicolumn{1}{c|}{5.0\%} & \multicolumn{1}{c|}{0.0\%} & \multicolumn{1}{c|}{0.0\%} & \multicolumn{1}{c|}{0.0\%} & \multicolumn{1}{c|}{92.8\%} \\ \hline
\multicolumn{1}{|l|}{ResNet-164} & \multicolumn{1}{c|}{29.0\%} & \multicolumn{1}{c|}{0.0\%} & \multicolumn{1}{c|}{5.4\%} & \multicolumn{1}{c|}{0.0\%} & \multicolumn{1}{c|}{0.0\%} & \multicolumn{1}{c|}{0.0\%} & \multicolumn{1}{c|}{93.8\%} \\ \hline
 & \multicolumn{7}{c}{CIFAR-100} \\ \cline{2-8} 
\multicolumn{1}{c|}{} & \multicolumn{1}{c|}{FGSM} & \multicolumn{1}{c|}{PGD} & \multicolumn{1}{c|}{BPDA} & \multicolumn{1}{c|}{MIM} & \multicolumn{1}{c|}{C\&W} & \multicolumn{1}{c|}{APGD} & \multicolumn{1}{c|}{Acc} \\ \hline
\multicolumn{1}{|l|}{ViT-B-32} & \multicolumn{1}{c|}{20.8\%} & \multicolumn{1}{c|}{1.9\%} & \multicolumn{1}{c|}{13.4\%} & \multicolumn{1}{c|}{3.1\%} & \multicolumn{1}{c|}{0.0\%} & \multicolumn{1}{c|}{0.0\%} & \multicolumn{1}{c|}{91.7\%} \\ \hline
\multicolumn{1}{|l|}{ViT-B-16} & \multicolumn{1}{c|}{20.4\%} & \multicolumn{1}{c|}{0.0\%} & \multicolumn{1}{c|}{11.9\%} & \multicolumn{1}{c|}{0.5\%} & \multicolumn{1}{c|}{0.0\%} & \multicolumn{1}{c|}{0.0\%} & \multicolumn{1}{c|}{92.8\%} \\ \hline
\multicolumn{1}{|l|}{ViT-L-16} & \multicolumn{1}{c|}{33.0\%} & \multicolumn{1}{c|}{1.6\%} & \multicolumn{1}{c|}{15.1\%} & \multicolumn{1}{c|}{4.7\%} & \multicolumn{1}{c|}{0.0\%} & \multicolumn{1}{c|}{0.0\%} & \multicolumn{1}{c|}{94.0\%} \\ \hline
\multicolumn{1}{|l|}{ViT-R50} & \multicolumn{1}{c|}{22.0\%} & \multicolumn{1}{c|}{0.2\%} & \multicolumn{1}{c|}{9.7\%} & \multicolumn{1}{c|}{0.4\%} & \multicolumn{1}{c|}{0.0\%} & \multicolumn{1}{c|}{0.0\%} & \multicolumn{1}{c|}{91.8\%} \\ \hline
\multicolumn{1}{|l|}{BiT-M-R50x1} & \multicolumn{1}{c|}{36.0\%} & \multicolumn{1}{c|}{0.0\%} & \multicolumn{1}{c|}{7.0\%} & \multicolumn{1}{c|}{0.0\%} & \multicolumn{1}{c|}{0.0\%} & \multicolumn{1}{c|}{0.0\%} & \multicolumn{1}{c|}{87.4\%} \\ \hline
\multicolumn{1}{|l|}{BiT-M-R101x3} & \multicolumn{1}{c|}{1.2\%} & \multicolumn{1}{c|}{0.0\%} & \multicolumn{1}{c|}{0.4\%} & \multicolumn{1}{c|}{0.0\%} & \multicolumn{1}{c|}{0.0\%} & \multicolumn{1}{c|}{0.0\%} & \multicolumn{1}{c|}{91.8\%} \\ \hline
\multicolumn{1}{|l|}{ResNet-56} & \multicolumn{1}{c|}{6.0\%} & \multicolumn{1}{c|}{0.2\%} & \multicolumn{1}{c|}{3.3\%} & \multicolumn{1}{c|}{0.4\%} & \multicolumn{1}{c|}{0.0\%} & \multicolumn{1}{c|}{0.0\%} & \multicolumn{1}{c|}{71.6\%} \\ \hline
\multicolumn{1}{|l|}{ResNet-164} & \multicolumn{1}{c|}{7.6\%} & \multicolumn{1}{c|}{0.3\%} & \multicolumn{1}{c|}{3.7\%} & \multicolumn{1}{c|}{0.9\%} & \multicolumn{1}{c|}{0.0\%} & \multicolumn{1}{c|}{0.0\%} & \multicolumn{1}{c|}{74.2\%} \\ \hline
 & \multicolumn{7}{c}{ImageNet} \\ \cline{2-8} 
\multicolumn{1}{l|}{} & \multicolumn{1}{c|}{FGSM} & \multicolumn{1}{c|}{PGD} & \multicolumn{1}{c|}{BPDA} & \multicolumn{1}{c|}{MIM} & \multicolumn{1}{c|}{C\&W} & \multicolumn{1}{c|}{APGD} & \multicolumn{1}{c|}{Acc} \\ \hline
\multicolumn{1}{|l|}{ViT-B-16} & \multicolumn{1}{c|}{23.1\%} & \multicolumn{1}{c|}{0.0\%} & \multicolumn{1}{c|}{7.3\%} & \multicolumn{1}{c|}{0.0\%} & \multicolumn{1}{c|}{0.0\%} & \multicolumn{1}{c|}{0.0\%} & \multicolumn{1}{c|}{80.3\%} \\ \hline
\multicolumn{1}{|l|}{ViT-L-16 (224)} & \multicolumn{1}{c|}{27.9\%} & \multicolumn{1}{c|}{0.0\%} & \multicolumn{1}{c|}{8.4\%} & \multicolumn{1}{c|}{0.0\%} & \multicolumn{1}{c|}{0.0\%} & \multicolumn{1}{c|}{0.0\%} & \multicolumn{1}{c|}{82.0\%} \\ \hline
\multicolumn{1}{|l|}{ViT-L-16 (512)} & \multicolumn{1}{c|}{29.8\%} & \multicolumn{1}{c|}{0.0\%} & \multicolumn{1}{c|}{8.4\%} & \multicolumn{1}{c|}{0.0\%} & \multicolumn{1}{c|}{0.0\%} & \multicolumn{1}{c|}{0.0\%} & \multicolumn{1}{c|}{85.4\%} \\ \hline
\multicolumn{1}{|l|}{BiT-M-R50x1} & \multicolumn{1}{c|}{28.7\%} & \multicolumn{1}{c|}{0.0\%} & \multicolumn{1}{c|}{3.5\%} & \multicolumn{1}{c|}{0.0\%} & \multicolumn{1}{c|}{0.0\%} & \multicolumn{1}{c|}{0.0\%} & \multicolumn{1}{c|}{79.9\%} \\ \hline
\multicolumn{1}{|l|}{BiT-M-R152x4} & \multicolumn{1}{c|}{60.9\%} & \multicolumn{1}{c|}{0.0\%} & \multicolumn{1}{c|}{15.2\%} & \multicolumn{1}{c|}{0.0\%} & \multicolumn{1}{c|}{0.0\%} & \multicolumn{1}{c|}{0.0\%} & \multicolumn{1}{c|}{85.3\%} \\ \hline
\multicolumn{1}{|l|}{ResNet-50} & \multicolumn{1}{c|}{11.8\%} & \multicolumn{1}{c|}{0.0\%} & \multicolumn{1}{c|}{1.4\%} & \multicolumn{1}{c|}{0.0\%} & \multicolumn{1}{c|}{0.0\%} & \multicolumn{1}{c|}{0.0\%} & \multicolumn{1}{c|}{74.5\%} \\ \hline
\multicolumn{1}{|l|}{ResNet-152} & \multicolumn{1}{c|}{18.1\%} & \multicolumn{1}{c|}{0.0\%} & \multicolumn{1}{c|}{2.7\%} & \multicolumn{1}{c|}{0.0\%} & \multicolumn{1}{c|}{0.0\%} & \multicolumn{1}{c|}{0.0\%} & \multicolumn{1}{c|}{77.0\%} \\ \hline
\end{tabular}
}
\end{table*}

\begin{table*}[]
\centering
\caption{White-box attack parameters for CIFAR-10.}
\begin{tabular}{|l|l|l}
\cline{1-2}
Attack & Parameters &  \\ \cline{1-2}
FGSM & $\epsilon$ = 0.031 &  \\ \cline{1-2}
PGD & $\epsilon$ = 0.031, $\epsilon_{step}$ = 0.00155,  steps = 20 &  \\ \cline{1-2}
BPDA & $\epsilon$ = 0.031, steps = 100,   max iterations = 100, learning rate = 0.5 &  \\ \cline{1-2}
MIM & $\epsilon$ = 0.031, $\epsilon_{step}$ = 0.00155, decay factor = 1.0 &  \\ \cline{1-2}
CW & confidence = 50, step size = 0.00155, steps = 30 &  \\ \cline{1-2}
APGD & $\epsilon$ = 0.031, number of restarts = 1, $\rho$ = 0.75,  $n^{2}$ queries = 5000 &  \\ \cline{1-2}
\end{tabular}
\end{table*}

\begin{table*}[]
\centering
\caption{White-box attack parameters for ImageNet.}
\begin{tabular}{|l|l|l}
\cline{1-2}
Attack & Parameters &  \\ \cline{1-2}
FGSM & $\epsilon$ = 0.062 &  \\ \cline{1-2}
PGD & $\epsilon$ = 0.062, $\epsilon_{step}$ = 0.0031,  steps = 20 &  \\ \cline{1-2}
BPDA & $\epsilon$ = 0.062, steps = 100,   max iterations = 100, learning rate = 0.5 &  \\ \cline{1-2}
MIM & $\epsilon$ = 0.062, $\epsilon_{step}$ = 0.0031, decay factor = 1.0 &  \\ \cline{1-2}
CW & confidence = 50, step size = 0.0031, steps = 30 &  \\ \cline{1-2}
APGD & $\epsilon$ = 0.062, number of restarts = 1, $\rho$ = 0.75,  $n^{2}$ queries = 5000 &  \\ \cline{1-2}
\end{tabular}
\end{table*}
\section{Self-Attention Gradient Attack (SAGA)}
\label{sup:saga}

In the main paper we introduced the Self-Attention Gradient Attack (SAGA). Here, we provide some additional experimental results and parameters related to our attack. 

\begin{figure*}
\centering
\includegraphics[scale=0.5]{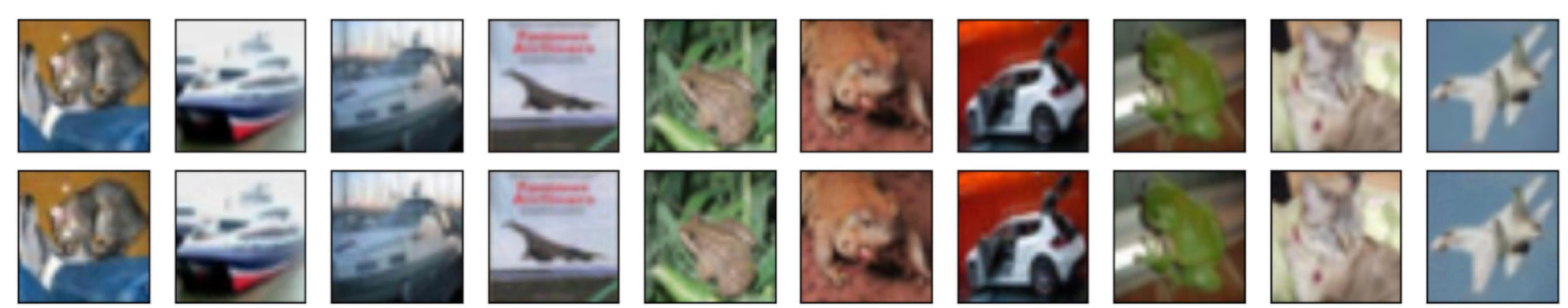}
\caption[]{Adversarial images generated using SAGA on CIFAR-10. The top row of images are the the clean images generated from the CIFAR-10 test set. The bottom row of images are the adversarial images generated using SAGA with the $l_{\infty}$ norm and $\epsilon=0.031$. These images correspond to SAGA when the models are ViT-L-16 and BiT-M-R101x3. Visually, there is very little perceivable difference between the clean and adversarial images generated by SAGA.}
 \label{fig:sagacifar}
 \noindent
\end{figure*}

\begin{figure*}
\centering
\includegraphics[scale=0.4]{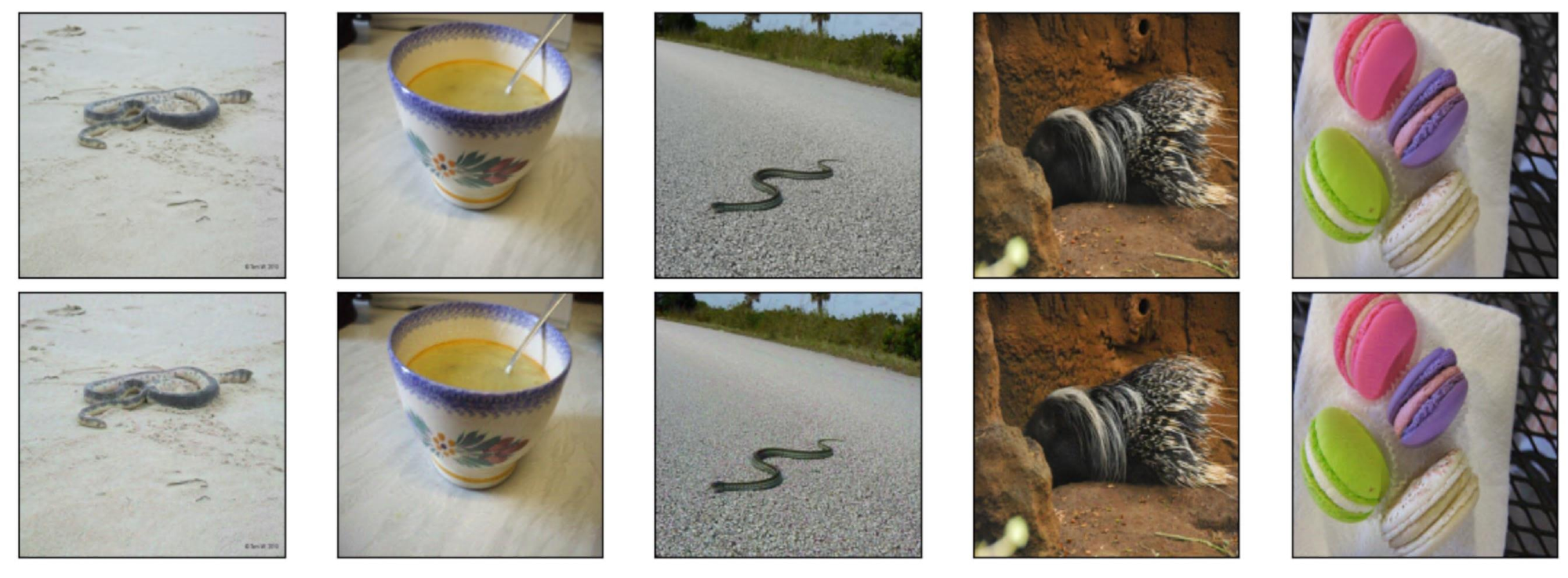}
\caption[]{Adversarial images generated using SAGA on ImageNet. The top row of images are the the clean images generated from the ImageNet validation set. The bottom row of images are the adversarial images generated using SAGA with the $l_{\infty}$ norm and $\epsilon=0.062$. These images correspond to SAGA when the models are ViT-L-16 and BiT-M-R152x4. Visually, there is very little perceivable difference between the clean and adversarial images generated by SAGA.}
 \label{fig:sagaImageNet}
 \noindent
\end{figure*}

\textbf{SAGA Adversarial Images:} In figure~\ref{fig:sagacifar} and figure~\ref{fig:sagaImageNet} we show some examples of the adversarial images generated by SAGA for CIFAR-10 and ImageNet. We generate these images from a defense comprised of ViT-L-16 and BiT-M-R101x3 for CIFAR-10 and ViT-L-16 and BiT-M-R152x4 for ImageNet. In the attack we use the $l_{\infty}$ norm and $\epsilon=0.031$ for CIFAR-10 and $\epsilon=0.062$ for ImageNet. From these figures, it is clear that just like other white-box attacks, SAGA is capable of creating adversarial examples with minimal visual perturbations. 

\textbf{SAGA Hyperparameters:} In the main paper we mentioned that the scaling factors $\alpha_{k}$ must be chosen carefully for each model when running SAGA. In table~\ref{table:sagasup} we give the $\alpha$ values and the corresponding robust accuracies for each attack. From the table, it can clearly be seen that simple averaging ($\alpha_{1}=0.5 \alpha_{2}=0.5$) does not yield a high attack success rate (low robust accuracy). For example, for the Bit/ViT defense for CIFAR-10 the robust accuracy with simple averaging yields an average robust accuracy of $47.5\%$. However, when each $\alpha$ is fine-tuned properly, it yields a robust accuracy of only $26\%$ (an attack success rate of $74\%$). 

It is also worth noting that some $\alpha$ values are very different in value. For example, $\alpha_{1}=0.998$ for ViT-L-16 but only $2e-4$ for BiT-M-R101x3 for SAGA for CIFAR-10. A natural question would be, why do we even need to take into account the gradient for BiT-M-R101x3, couldn't it be $0$? The simple answer to this question is even though sometimes the $\alpha$ values are small, they are critical to crafting adversarial examples that are misclassified by BOTH models. For empirical proof that a single gradient does not suffice, one needs only to look again at the transferability results in table~\ref{table:transfer} in the main paper. 

\textbf{ResNet SAGA Results:} In the main paper we tested Vision Transformer/Big Transfer Model combinations. SAGA can also can work on other Vision Transformer/CNN combinations as well. In table~\ref{table:sagasup} we demonstrate a proof of concept of this by attacking a ViT-L-16/ResNet-164 pair for CIFAR-10. Here we can see that just like that ViT/BiT combination, the ViT/ResNet combination is not secure against SAGA as the robust accuracy is only $15\%$.

\begin{table*}[]
\centering
\caption{Self-Attention Gradient Attack (SAGA) results for CIFAR-10, CIFAR-100 and ImageNet. In the table $\alpha_{1}$ represents the coefficient used to scale the gradient of the ViT model and $\alpha_{2}$ represents the coefficient used to scale the gradient of the respective CNN. In the table ViT corresponds to ViT-L-16 and BiT corresponds to BiT-M-R101x3 for CIFAR-10 and CIFAR-100 and BiT-M-R152x4 for ImageNet. ResNet corresponds to ResNet-164.}
\label{table:sagasup}
\begin{tabular}{l|c|c|c|c|c|}
\cline{2-6}
 & \multicolumn{5}{c|}{CIFAR-10} \\ \cline{2-6} 
 & \multicolumn{1}{l|}{$\alpha_{1}$} & \multicolumn{1}{l|}{$\alpha_{2}$} & \multicolumn{1}{l|}{Robust Acc ViT} & \multicolumn{1}{l|}{Robust Acc CNN} & \multicolumn{1}{l|}{Average Robust Acc} \\ \hline
\multicolumn{1}{|l|}{ViT/BiT} & 0.5 & 0.5 & 94.9\% & 0.1\% & 47.5\% \\ \hline
\multicolumn{1}{|l|}{ViT/BiT} & 0.9998 & 2.00E-04 & 27.3\% & 24.7\% & 26.0\% \\ \hline
\multicolumn{1}{|l|}{ViT/ResNet} & 0.5 & 0.5 & 7.3\% & 38.3\% & 22.8\% \\ \hline
\multicolumn{1}{|l|}{ViT/ResNet} & 0.01 & 0.99 & 15.1\% & 14.8\% & 15.0\% \\ \hline
 & \multicolumn{5}{c|}{CIFAR-100} \\ \cline{2-6} 
 & $\alpha_{1}$ & $\alpha_{2}$ & Robust Acc ViT & Robust Acc CNN & Average Robust Acc \\ \hline
\multicolumn{1}{|l|}{ViT/BiT} & 0.5 & 0.5 & 3.7\% & 48.9\% & 26.3\% \\ \hline
\multicolumn{1}{|l|}{ViT/BiT} & 0.9985 & 0.0015 & 16.7\% & 14.5\% & 15.6\% \\ \hline
 & \multicolumn{5}{c|}{ImageNet} \\ \cline{2-6} 
 & $\alpha_{1}$ & $\alpha_{2}$ & Robust Acc ViT & Robust Acc CNN & Average Robust Acc \\ \hline
\multicolumn{1}{|l|}{ViT/BiT} & 0.5 & 0.5 & 56.7\% & 0.2\% & 28.5\% \\ \hline
\multicolumn{1}{|l|}{ViT/BiT} & 0.99 & 0.01 & 13.3\% & 12.0\% & 12.7\% \\ \hline
\end{tabular}
\end{table*}

\section{Black-Box Attacks}
\label{sup:blackbox}

In this section, we mathematically define the black-box adversarial model. We also give detailed descriptions of the black-box attacks tested in this paper. Along with the attack descriptions, we also list the parameters we choose for each attack. Unlike section~\ref{sup:whitebox} where a single adversarial model suffices, here we break the black-box adversarial model into two distinct types: query based and transfer based models.  

Before we discuss the differences, it is important to note the basic commonality between the two threat models. A successful attack is still defined the same for all threat models. Specifically, the three conditions we previously defined must hold. First, the adversarial sample must be misclassified (equation~\ref{eq:condition}). Second, the adversarial sample $x_{adv}$ must be within a certain distance of the original sample $x$ (equation~\ref{eq:condition2}). Third, the adversarial sample must have pixels within a valid range (equation~\ref{eq:condition3}).

The other commonality between the two black-box adversarial models are the components that make up the defense: a classifier $C$ with trained parameters $\theta$. In contrast to the white-box adversary, we will also explicitly define additional training components. We define the training samples that $C$ was trained on to obtain $\theta$ as the set $(X,Y)$. Let us also define the pre-training dataset as $(X^{'},Y^{'})$. Here the pre-training dataset is only applicable to Vision Transformers and Big Transfer Models where the pre-training dataset $(X^{'},Y^{'})$ is ImageNet-21K and the training dataset $(X,Y)$ is either CIFAR-10, CIFAR-100 or ImageNet. 

\subsection{Query Based Adversarial Model}

\textbf{Adversarial Capabilities:} For the query based adversarial model the attacker lacks knowledge of $\theta$, the specific classifier architecture $C$, the training set $(X,Y)$ and the pre-training set $(X^{'},Y^{'})$. The adversary starts with a  clean example $x$ and is able to query the classifier $C$, with different perturbations (e.g. $x+\epsilon$).The adversarial model here is constrained by the fact that for each example $x$, only a fixed number of queries $q$ can be done on $C$. In this threat model the type of response from the classifier $C$ also matters. When the adversary queries $C$, the defense can return either the hard label (class label only) or the corresponding probability vector. To narrow our scope of study, in this paper we consider the adversary which only has access to the hard label. 

\textbf{Attack Setup and Discussion:} To test query based adversaries we use the RayS attack~\cite{chen2020rays}. We use $1000$ clean examples for CIFAR-10 and ImageNet. In our attacks we set the query budget $q$ to be $10,000$ for each sample. We use $\epsilon=0.031$ for CIFAR-10 and $\epsilon=0.062$ for ImageNet in conjunction with the $l_{\infty}$ norm. Due to the high computational complexity of the attack, we only test single models for CIFAR-10 and ImageNet. We omit CIFAR-100 and BiT-M-R152x4. Our attack results are shown in table~\ref{table:rayS}. In general, it can be seen single models are not robust to the RayS attack as no model has more than $30\%$ robust accuracy. 

\begin{table*}[]
\centering
\caption{RayS attack on single classifiers for CIFAR-10 and ImageNet. The robust accuracy for each model is reported in the table.}
\label{table:rayS}
\begin{tabular}{l|c|}
\cline{2-2}
 & \multicolumn{1}{l|}{RayS CIFAR-10} \\ \hline
\multicolumn{1}{|l|}{ResNet-56} & 0.8\% \\ \hline
\multicolumn{1}{|l|}{ResNet-164} & 0.0\% \\ \hline
\multicolumn{1}{|l|}{ViT-B-16} & 8.2\% \\ \hline
\multicolumn{1}{|l|}{ViT-B-32} & 11.1\% \\ \hline
\multicolumn{1}{|l|}{ViT-L-16} & 14.5\% \\ \hline
\multicolumn{1}{|l|}{R50-ViT-B-16} & 22.9\% \\ \hline
\multicolumn{1}{|l|}{BiT-M-R50x1} & 0.9\% \\ \hline
\multicolumn{1}{|l|}{BiT-M-R101x3} & 3.7\% \\ \hline
 & \multicolumn{1}{l|}{RayS ImageNet} \\ \hline
\multicolumn{1}{|l|}{ResNet-50} & 3.1\% \\ \hline
\multicolumn{1}{|l|}{ResNet-152} & 2.7\% \\ \hline
\multicolumn{1}{|l|}{ViT-B-16-224} & 1.6\% \\ \hline
\multicolumn{1}{|l|}{ViT-L-16} & 25.9\% \\ \hline
\multicolumn{1}{|l|}{ViT-L-16-224} & 3.3\% \\ \hline
\multicolumn{1}{|l|}{BiT-M-R50x1} & 3.1\% \\ \hline
\end{tabular}
\end{table*}

\subsection{Transfer Based Adversarial Model}

\textbf{Adversarial Capabilities:} The transfer based adversary is granted a wide range of abilities. Specifically, a transfer based adversary may know part or all of the original training data $(X,Y)$ for $C$. They may also have access to the pre-training data $(X^{'},Y^{'})$. Unlike query based adversaries, the transfer based adversary is not restricted in terms of the number of queries made to $C$. The only thing not given to the adversary is the architecture classifier for $C$ and the trained parameters $\theta$. The general strategy for the transfer based adversary is as follows: the attacker starts with an untrained classifier $S$. Note $S$ is often referred to as the synthetic model. If the adversary has access to the pre-training data, they start by training $S$ with $(X^{'},Y^{'})$. The adversary then queries $C$ to label the training set $X$. They then trains $S$ on $(X,\hat{Y})$, where $\hat{Y}$ are the hard class labels obtained from $C$. Once $S$ has been trained, a white-box attack $A$ can be run on $S$ to generate adversarial examples. These examples are then applied to $C$ in the hopes that the adversarial samples are able to "transfer" from $S$ to $C$.

\begin{figure*}
\centering
\includegraphics[scale=0.65]{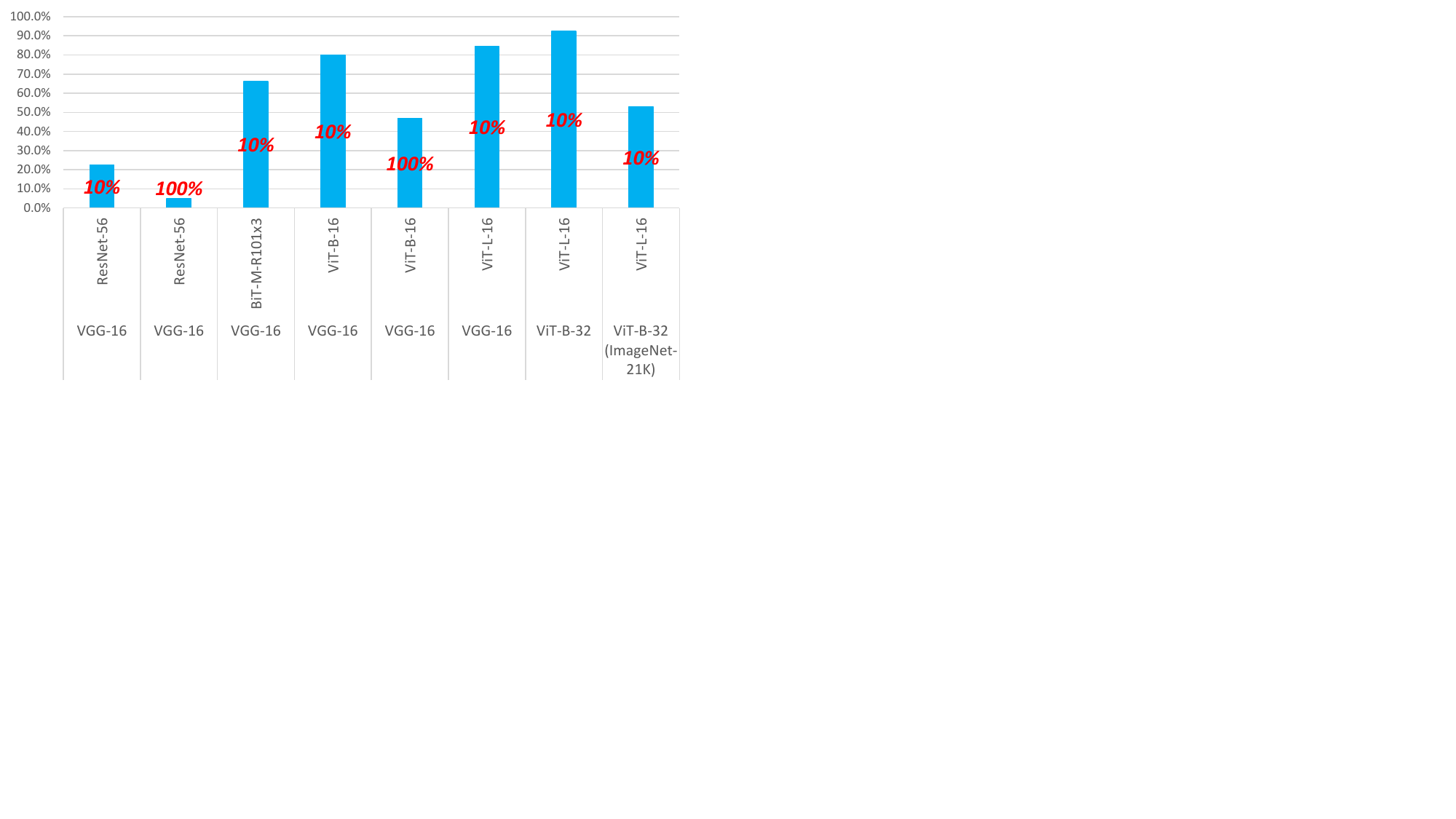}
\caption[]{Black-box transfer attack hyperparameter experimental results for CIFAR-10. The text in red on top of the bars indicates the strength of the attack (what percent of the training data is available to the adversary). The bars themselves represent the robust accuracy of the model under attack. The horizontal text represents the type of synthetic model $S$, used in the attack. The  vertical text represents the model being attacked. This experiment is important because it indicates which type of synthetic model works best when doing the actual attack.}
 \label{fig:hyperparam}
 \noindent
\end{figure*}

\textbf{Attack Setup and Discussion:} For a transfer attack several components must be selected. These components include the synthetic architecture $S$, the percent of training data $(X,Y)$ visible to the adversary, and the type of white-box attack $A$ that will be used on $S$ to generate adversarial examples. Ideally, we want to test under the strongest possible adversary. This means a careful choice of $S$ and using $100\%$ of the training data. However, as these experiments are time consuming (each attack requires training a synthetic model from scratch) we first conduct several smaller scale experiments to help us choose the hyperparameters for the main attack. These results are show in table~\ref{table:hyper} and figure~\ref{fig:hyperparam}. For $A$ we use MIM attack to generate samples with $S$. We set the maximum perturbation $\epsilon=0.031$ for CIFAR-10 and experiment with a range of different synthetic models $S$.

From the hyperparameter experiments we can notice a few interesting results. First of all when attacking a Vision Transformer like ViT-L-16, the choice of synthetic model greatly effects the robust accuracy. Even when only $10\%$ of the data is available if $S$ is a Vision Transformer (ViT-B-32) and it is pre-trained on ImageNet-21K, the robust accuracy is only of ViT-L-16 is only $53\%$. If the attacker uses a synthetic model that is NOT pre-trained (but still ViT-B-32) the robust accuracy is $92.4\%$. Now this brings up a new dilemma for the attacker. Originally, in attacks on CNNs the architecture of the synthetic model did not greatly effect the performance of the attack~\cite{papernot2017practical}. Likely this is in part do the fact that these attacks were transferring samples from CNNs to other CNNs. However, the same result doesn't hold for Vision Transformers. Using a CNN (like VGG-16) does not give a very high attack success rate. We can see that when we do a $100\%$ strength attack on ViT-L-16 using VGG-16, the robust accuracy is still $46.8\%$. Compare this result to the same attack with ResNet-56 and it can be seen that the robust accuracy is only $4.8\%$.

In our hyperparameter experiments our main goal was to find a good set of parameters for attacking Vision Transformer based defenses. In that respect, our experiments accomplish this. We can see using a pre-trained ViT-B-32 with even just $10\%$ of the training data gives good attack results. However, our experiments also uncover a much more interesting concept. Unlike with CNNs where the choice of architecture was not critical in the transfer attack, initial experiments show Vision Transformers mandate a careful choice of synthetic model $S$. By merely using Vision Transformers in a defense, the transfer based attacker is put at a new disadvantage. While it is beyond the scope of this work to explore this notion further, it does pose an interesting new challenge for black-box attack designers. 

\begin{table*}[]
\label{table:hyper}
\caption{Results of CIFAR-10 hyperparameter experiments for transfer attacks using different strength attacks and different synthetic models.}
\centering
\begin{tabular}{|l|l|c|c|}
\hline
Defense Model & Synthetic Model & \multicolumn{1}{l|}{Attack Strength} & \multicolumn{1}{l|}{Robust Acc} \\ \hline
ViT-B-16 & VGG-16 & 10.0\% & 79.9\% \\ \hline
ViT-B-16 & VGG-16 & 100.0\% & 46.8\% \\ \hline
ViT-L-16 & VGG-16 & 10.0\% & 84.4\% \\ \hline
ViT-L-16 & ViT-B-32 & 10.0\% & 92.4\% \\ \hline
ViT-L-16 & ViT-B-32 (ImageNet-21K) & 10.0\% & 53.0\% \\ \hline
ResNet-56 & VGG-16 & 10.0\% & 22.6\% \\ \hline
ResNet-56 & VGG-16 & 100.0\% & 4.8\% \\ \hline
BiT-M-R101x3 & VGG-16 & 10.0\% & 66.1\% \\ \hline
\end{tabular}
\end{table*}

\section{Discussion on White-Box and Transfer Attacks on the Vision Transformer}
\label{sup:whitebox_transformer}

The Vision transformer as reported in~\cite{dosovitskiy2021an} is an encoder-based architecture. This architecture is an adaptation of the Transformer architecture popular in Natural Language Processing applications. While the original Transformer from Vaswani et al.~\cite{vaswani2017attention} used both encoders and decoders for sequence-to-sequence applications, the Vision transformer is only Encoder-based. The input image to be fed to the transformer is divided into equal-size patches. Then the sequence of these patches passes through an embedding layer. Positional encoding is added to the embedding vector feeding to a layer of encoders. The output dimension of each encoder matches the input dimension, which makes it easy to stack these encoders. The output from the last encoder is fed to a linear network layer acting as a classifier.

The building blocks of an encoder are the “Attention” network, followed by Batch Normalization with skip connections. Many attention blocks are used in parallel (similar to feature maps in CNNs), which are referred to as the “multi-headed self-attention network”. The self-attention block uses three linear networks of query, key and value parametrized by $W_k$, $W_q$, and $W_v$ matrices. The query and the key correspond to the positions of the input patches in image $X$ for which we are interested in computing the attention with respect to each other. Once we have computed the self-attention of the entire input set, we turn it into a probability distribution by the Softmax function. An encoder uses multiple attention computations in parallel, where each attention block is referred to as an attention head. 
Let $N$ be the number of $p $ x $ p$ patches in an $n$ x $m$ image, and $e$ be the embedding size for each patch, then the positionally encoded input to the encoder is:
 \begin{equation}
    \label{eq:tran1}
    X_p=X+P	  
    \end{equation}
where $X$, $P\in \mathbb{R}^{N %\text{x} 
\times e}$, and $P$ is the positional encoding for the image patches.
The computation for each attention head $i$ in terms of key, query and value networks can be described as:
\begin{align} \label{eq:tran2}
    K_i=X_p W_{k,i} 	\\
Q_i=X_p W_{q,i}			\\
V_i=X_p W_{v,i}			
    \end{align}
The self-attention $A_i$ in an attention head $i$ with $n_h$ number of heads is computed as:
\begin{equation}
    \label{eq:tran3}
    A_i=[softmax(Q_i K_i^T )/n_h]V_i	
    \end{equation}
The output from all attention heads is concatenated and passed through a linear layer parameterized by $W_o$, as shown below:
\begin{equation}
    \label{eq:tran4}
    M_A=[concat_{i \in n_{h}}[A_i]]W_o+b_o				
    \end{equation}
	
In a transformer, the multi-headed self-attention passes through a batch normalization layer with the input being added to the output of batch normalization in a ResNet-like manner. Then, it further passes through a linear, and another batch normalization layer. Thus, the output of the first encoder $Enc$ as a function of the position encoded input, $X_p$  , can be described as:
    \begin{multline} 
    \label{eq:tran5}
     Enc(X_p)=batchnorm([[batchnorm(M_A )+ \\X_p] W_l+b_l ])+[batchnorm(M_A )+ X_p ]		
    \end{multline}
where $W_l$ and $b_l$ are the parameters of the linear network of the encoder. After passing the input through a series of encoders, a classifier is connected to the last output of the final encoder, as:
\begin{equation}
\label{eq:tran6}
    y=Classifier(Enc(\ldots Enc(X_p)\ldots ))		 
\end{equation}

For any differentiable loss function $L$, operating on the output of the transformer as given by equation~\ref{eq:tran6}, the computation of $\partial L / \partial x = \nabla_x (L)$ requires that all components of equation~\ref{eq:tran6} be differentiable with respect to the input $x$. Since equations~\ref{eq:tran2} -~\ref{eq:tran5} that contribute to equation~\ref{eq:tran6} including the batch normalization are all closed-form differentiable with respect to position encoded input $X_p$,  and since $X_p$ is a simple function of $x$,  $\nabla_x (L)$ can be computed.  Thus an adversarial image can be efficiently created using a white-box attack formulation. We confirm this empirically, as all white-box attacks successfully compromise the classifier accuracies, resulting in zero robust accuracy for many white-box attacks.

\subsection{Transfer Attacks}From a black box adversarial robustness point of view, one category of attacks is referred to as transfer attacks. Here a new network model is created (referred to as the synthetic model) and trained either on the same dataset as the model under attack, or creating training data from querying the input-output behavior of the target model. One fundamental question to ask is how the transformer-based models behave with respect to transfer attacks from a CNN-based synthetic model and vice versa. An equivalency of the Transformer model with the CNN based models under some simplified assumptions was presented in~\cite{Cordonnier2020On}. The multi-headed attention at pixel $q$ for attention head $h$ is expressed as (this is similar to our development in equation ~\ref{eq:tran4}):
\begin{equation}
\label{eq:tran7}
    M_A(X)_{q_{,:}} = \sum_{h \in[N_h]} (\sum_{k}softmax(A_{q,:}^h)_k X_{k_{,:}})W^{(h)} + b_{out}	 
\end{equation}
For the $h$-th attention head, the attention probability is one when $k = q-f(h)$ and zero otherwise. The layer's output at pixel $q$ is then shown to be equal to:
\begin{equation}
\label{eq:tran8}
    M_A(X)_q = \sum_{h \in[N_h]} X_{q-f(h),:}W^{(h)} + b_{out}	 
\end{equation}
The above can be seen to be equivalent to the convolution operation. The development in~\cite{Cordonnier2020On} as shown above demonstrates an equivalence in the Transformer and the CNN. Here we mean equivalence in the sense that the transformer can equivalently perform a $k\times k$ convolution if an appropriate value matrix $W_v$ is chosen in an attention head. Whether a transformer actually learns the appropriate $W_v$ to perform the equivalent convolution can only be answered empirically. Part of this empirical answer is given in the following subsection where we study the decision regions created by Transformers and CNNs.  

%Kaleel, your fish graphs showing boundary effects in transformers and Resnets and the corresponding explanation will go well here.

\subsection{Decision Region Graphs}

One way to visually comprehend the transferability between different models is to examine their decision region graphs. A decision region graph can be a visual representation of the different classification regions of a model using a color coded 2-D graph. Decision region graphs for CNNs trained on ImageNet were originally shown in~\cite{LiuTransfer2017}. 

For every dataset and model in this paper, we construct the decision region graph. Formally, we can describe the generation of the graph as follows: Each graph is constructed with respect to a single image $I$. For every model, we use the same image $I$ to build the graph (i.e. we use sample 49443 from the validation set of ImageNet). Every point on the graph corresponds to a class label for the given image. The origin $(x=0,y=0)$ corresponds to the original (unperturbed) image. Outside the origin, the image is perturbed according to the following equation:

\begin{equation}
\label{eq:fish}
I'=I+x \cdot g+y \cdot r
\end{equation}
where $I'$ represents the new perturbed image, $I$ represents the original image, $g$ represents the gradient of the image with respect to the loss function of the model, and $r$ represents a random noise orthogonal to $g$. In equation~\ref{eq:fish}, $x$ and $y$ represent coordinates on the graph which control the magnitude of the adversarial noise $g$ and random noise $r$.

\begin{figure*}
\centering
\includegraphics[scale=0.5]{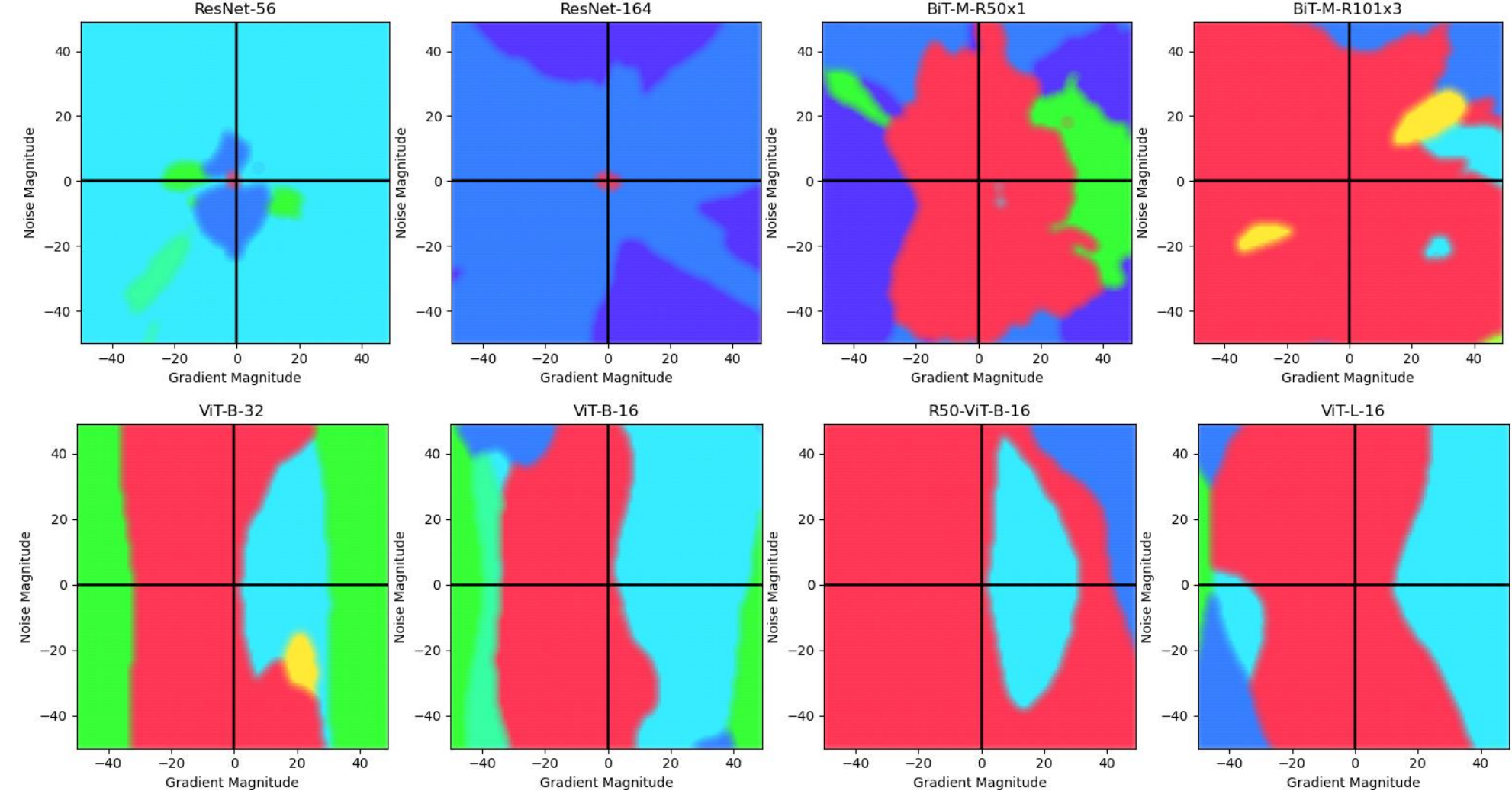}
\caption[]{Vision Transformer, Big Transfer Model and ResNet decision regions for CIFAR-10.}
 \label{fig:cifar10Fish}
 \noindent
\end{figure*}

\begin{figure*}
\centering
\includegraphics[scale=0.5]{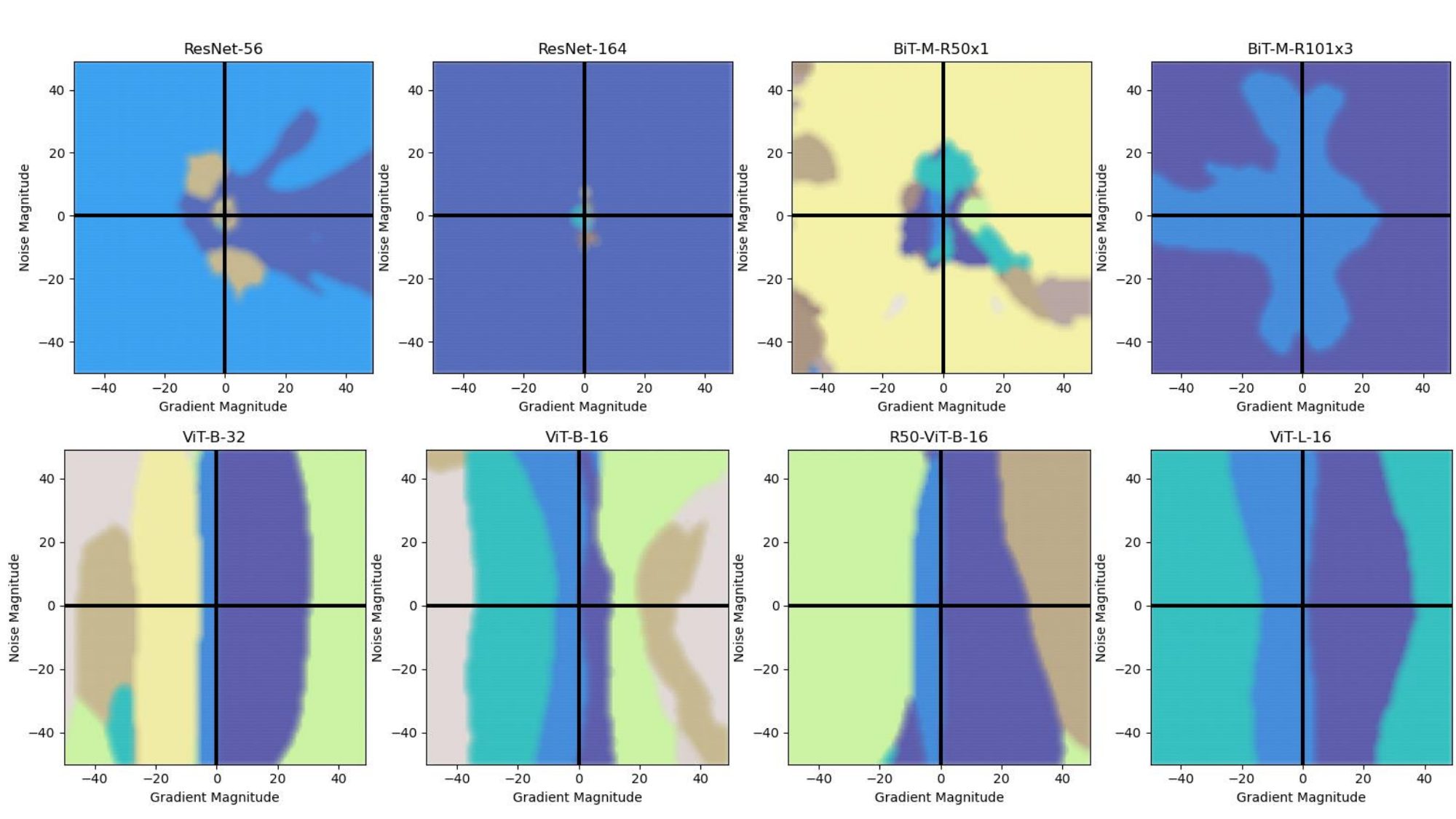}
\caption[]{Vision Transformer, Big Transfer Model and ResNet decision regions for CIFAR-100.}
 \label{fig:cifar100Fish}
 \noindent
\end{figure*}

\begin{figure*}
\centering
\includegraphics[scale=0.5]{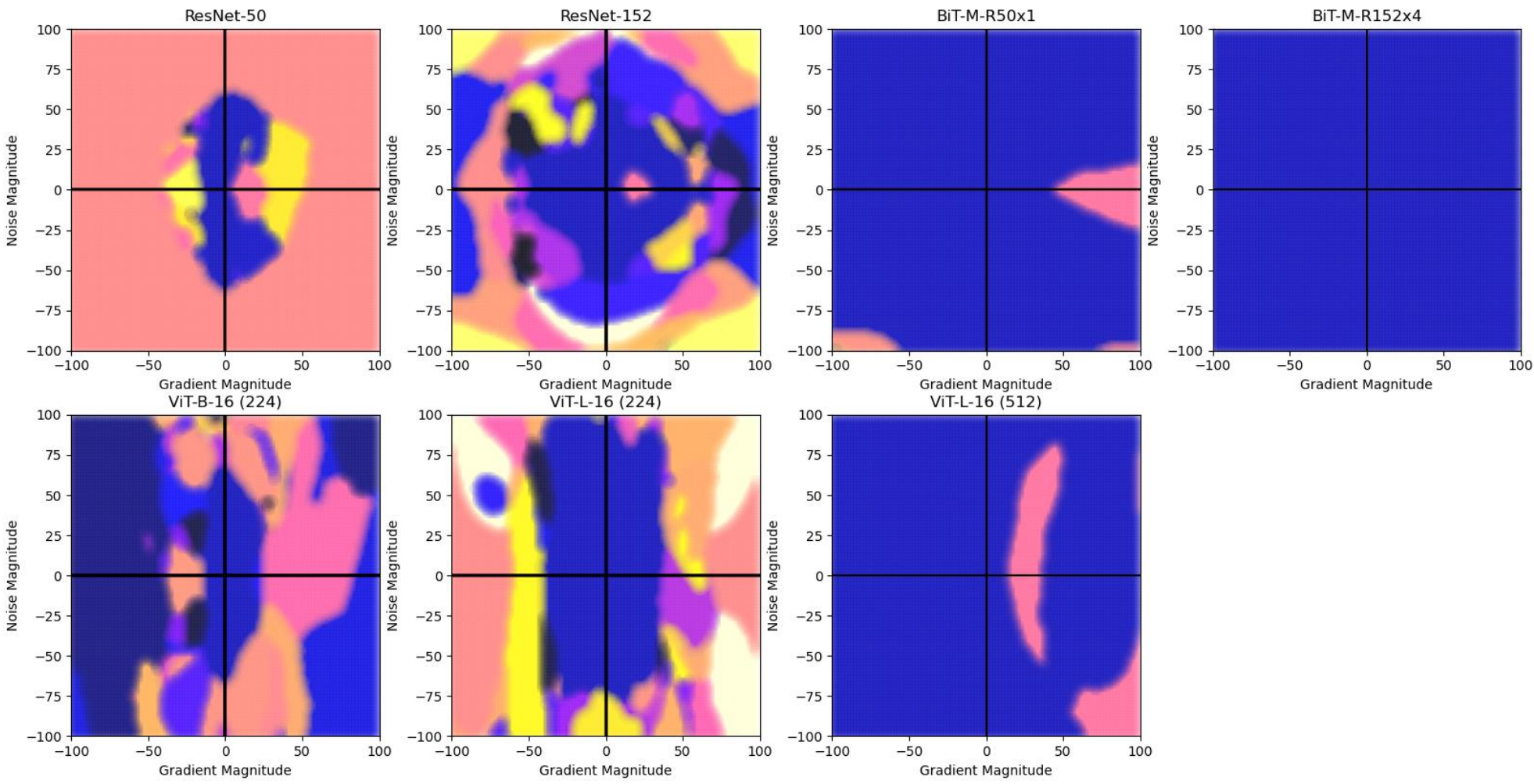}
\caption[]{Vision Transformer, Big Transfer Model and ResNet decision regions for ImageNet.}
 \label{fig:imageNetFish}
 \noindent
\end{figure*}

The decision region graphs may be slightly difficult to grasp at first but it comes with a natural intuitive explanation. The origin of the graph represents the unperturbed image $I$ with the correct class label. As we move in the $x$ direction on the graph, we increase the magnitude of the adversarial noise $g$ that is added to $I$. This is analogous to an FGSM attack using $I$ in which we keep increasing the size of the step ($\epsilon$ in equation~\ref{eq:fgsm}). As we move in the $y$ direction on the graph, this represents adding more and more random noise to $I$. When we move in both the $x$ and $y$ directions, it represents a combination of adding random noise and adversarial noise to the image. The last component of the graph, color, represents the class label that the model produces based on the perturbed input $I'$. Essentially, a decision region graph gives intuition about how the model classifies images that are noisy and adversarial. The decision region graphs for CIFAR-10, CIFAR-100 and ImageNet are shown in figures~\ref{fig:cifar10Fish},~\ref{fig:cifar100Fish} and~\ref{fig:imageNetFish}.

\textbf{Decision Region Graph Analysis:} In figure~\ref{fig:cifar10Fish}, the correct class label is represented by the color red. As we can see at the origin, all models correctly classify the sample. It can be noted that for the ResNets (ResNet-56 and ResNet-164), their robustness is quite limited. We can see for both these models there is only a small red sphere around the origin. As we move to larger and larger perturbations, the image quickly becomes misclassified (the blue regions). For the Vision Transformers and Big Transfer Models, we can see that they are much more tolerant of noise. For example, if we consider moving along the $y$ axis (adding random noise), none of the Vision Transformers misclassify the image. 

For figures~\ref{fig:cifar100Fish} and~\ref{fig:imageNetFish}, we can see a similar trend applies. In figure~\ref{fig:cifar100Fish}, light blue represents the correct class label and in figure~\ref{fig:imageNetFish}, dark blue represents the correct class label. In general, for both these figures we see the Vision Transformers tend to handle random noise well (see along the $y$ axis) and the ResNets are very sensitive to perturbations. It should also be noted in figure~\ref{fig:imageNetFish}, the graph for BiT-M-R152x4 is completely dark blue. This means that despite large perturbations, the model never fails to correctly classify the image $I$. This should not be completely surprising as BiT-M-R152x4 is one of the most complex models (in terms of number of parameters) that we experiment with. We mention complexity because it has been previously noted that model complexity alone helps thwart adversarial attacks~\cite{PGDmadry2018}. 

There is one other important take away, the landscape of the decision regions themselves are very different between model genuses. In figure~\ref{fig:cifar10Fish}, for the ResNet models we can see a small sphere of red surrounded by blue, for the Vision Transformers we can see large red regions around the $y$ axis and large light blue regions as we move in the $x$ direction. While we cannot directly make conjectures based on visualizations, the graphs do tend to support our main findings. Specifically, we know from the results in table~\ref{table:transfer} that the transferability between different models genuses is low. The decision region graphs lend credence to this claim by visually showing that the decisions regions between Vision Transformers, ResNets and Big Transfer Models do indeed look very different. Thus, we conjecture that even though a Vision Transformer is capable of implementing convolutions as described in~\cite{Cordonnier2020On}, in practice we observe that this may not be the case due to differing patterns of decision boundaries. 

\section{Friendly Adversarial Training Defense for Vision Transformers}
\label{sup:FAT_defense}

Friendly Adversarial Training (FAT)~\cite{zhang2020attacks} was recently proposed to improve the adversarial defense of deep networks. It is a simple training technique that uses less strong adversarial examples by employing an early stopping of the PGD algorithm. By incorporating another parameter, $\tau$, in the PGD $k$-step algorithm (referred to as PGD-$K$-$\tau$), the step amount $\tau$ by which the adversarial example crosses the decision boundary can be easily controlled.
The pseudocode for the FAT algorithm~\cite{zhang2020attacks} is described below:
\begin{align*} \label{eq:tran2}
& \mathbf{while}\;K > 0\; \mathbf{do}	\\
&\;\;\;\;\mathbf{if}\;\text{arg max}_{i}\;f(\widetilde{x}) \neq y\; \text{and}\;\tau = 0\; \mathbf{then}			\\
&\;\;\;\;\;\;\;\;\;\mathbf{break}	\\
&\;\;\;\;\mathbf{else\;if}\;\text{arg max}_{i}\;f(\widetilde{x}) \neq y\; \mathbf{then} \\
&\;\;\;\;\;\;\;\;\;\tau \xleftarrow\;\tau - 1	\\
&\;\;\;\;\mathbf{end\;if}	\\
&\;\;\;\;\widetilde{x} \xleftarrow\; P(\alpha\;\text{sign} (\nabla_{\widetilde{x}} l (f(\widetilde{x}), y)) + \widetilde{x})	\\
&\;\;\;\;K \xleftarrow\; K - 1   	\\ 
& \mathbf{end \;while}
    \end{align*}
    
As can be easily seen from the pseudocode, if $K =\tau$, the PGD-$K$-$\tau$ algorithm becomes equivalent to the standard PGD $k$-step algorithm. The application of FAT for creating adversarial defense for ResNets had a slightly better robust accuracy than e.g., Madry training, and a relatively lower drop in clean accuracy~\cite{zhang2020attacks}. For example, with FAT on the WRN-30-10, the clean accuracy dropped from around 95$\%$ to around 89$\%$, resulting in a robust accuracy of around 46$\%$ when the PGD-20 attack was used.

We evaluate the performance of different Transformer-based networks for different values of $\tau$. The results are presented in Table~\ref{table:fat_cifar}. From Table~\ref{table:fat_cifar}, it can be seen that the adversarial robustness of the Vision Transformers with respect to FAT is similar to the ResNet-based architectures. For example, for ViT-L-16 with $\tau = 1$, the clean accuracy drops from 99.1$\%$ to 94.2$\%$. The adversarial robustness in this case is approximately 47$\%$.

\begin{table*}[]
\caption{FAT defense accuracy for ViT-B-32, ViT-B-16, ViT-L-16, and ResNet-164 architectures on CIFAR-10 and CIFAR-100. The leftmost column in the table lists the model being tested; each model includes a subset of $\tau$ parameters for $\tau=0, 1, 2, 10$. The top row in the table lists the attacks run, from FGSM to APGD. The last column in the table lists the clean accuracy of the tested model.}
\centering
\label{table:fat_cifar}
\begin{tabular}{|l|c|c|c|c|c|c|c|}
\hline
 & \multicolumn{7}{c|}{CIFAR-10} \\ \hline
 & FGSM & PGD & BPDA & MIM & C\&W & APGD & Acc \\ \hline
ViT-B-32 & 37.9\% & 1.8\% & 17.6\% & 4.4\% & 0.0\% & 0.0\% & 98.6\% \\ \hline
$\tau = 0$ & 30.8\% & 17.5\% & 14.5\% & 17.3\% & 1.5\% & 1.5\% & 95.5\% \\ \hline
$\tau = 1$ & 33.9\% & 32.2\% & 16.4\% & 23.6\% & 9.7\% & 5.9\% & 93.2\% \\ \hline
$\tau = 2$ & 37.8\% & 40.3\% & 23.6\% & 31.4\% & 20.8\% & 13.9\% & 90.8\% \\ \hline
$\tau = 10$ & 42.6\% & 51.1\% & 33.3\% & 38.8\% & 34.1\% & 29.4\% & 78.9\% \\ \hline
 & \multicolumn{1}{l|}{} & \multicolumn{1}{l|}{} & \multicolumn{1}{l|}{} & \multicolumn{1}{l|}{} & \multicolumn{1}{l|}{} & \multicolumn{1}{l|}{} & \multicolumn{1}{l|}{} \\ \hline
ViT-B-16 & 39.5\% & 0.0\% & 20.3\% & 0.3\% & 0.0\% & 0.0\% & 98.9\% \\ \hline
$\tau = 0$ & 42.3\% & 34.0\% & 19.2\% & 29.0\% & 9.4\% & 4.1\% & 95.9\% \\ \hline
$\tau = 1$ & 43.2\% & 41.1\% & 26.3\% & 33.7\% & 19.3\% & 13.7\% & 93.8\% \\ \hline
$\tau = 2$ & 62.3\% & 25.4\% & 33.9\% & 25.1\% & 6.7\% & 5.4\% & 93.8\% \\ \hline
$\tau = 10$ & 43.1\% & 52.3\% & 36.8\% & 40.2\% & 35.0\% & 33.7\% & 73.3\% \\ \hline
 &  &  &  &  &  &  &  \\ \hline
ViT-L-16 & 56.3\% & 1.2\% & 28.70\% & 5.9\% & 0.0\% & 0.0\% & 99.1\% \\ \hline
$\tau = 0$ & 51.7\% & 43.6\% & 29.2\% & 39.3\% & 20.6\% & 15.4\% & 95.7\% \\ \hline
$\tau = 1$ & 49.1\% & 47.0\% & 31.9\% & 39.4\% & 26.8\% & 19.6\% & 94.2\% \\ \hline
$\tau = 2$ & 57.4\% & 48.8\% & 33.5\% & 40.2\% & 29.4\% & 21.8\% & 92.4\% \\ \hline
$\tau = 10$ & 49.5\% & 55.4\% & 33.7\% & 45.8\% & 37.7\% & 33.6\% & 85.3\% \\ \hline
 & \multicolumn{1}{l|}{} & \multicolumn{1}{l|}{} & \multicolumn{1}{l|}{} & \multicolumn{1}{l|}{} & \multicolumn{1}{l|}{} & \multicolumn{1}{l|}{} & \multicolumn{1}{l|}{} \\ \hline
ResNet-164 & 14.4\% & 3.0\% & 9.0\% & 2.2\% & 0.1\% & 0.0\% & 93.2\% \\ \hline
$\tau = 0$ & 47.7\% & 50.8\% & 39.5\% & 42.5\% & 34.6\% & 27.0\% & 90.3\% \\ \hline
$\tau = 1$ & 53.0\% & 56.2\% & 47.2\% & 49.0\% & 42.7\% & 34.4\% & 88.0\% \\ \hline
$\tau = 2$ & 56.2\% & 61.3\% & 50.0\% & 51.9\% & 46.9\% & 37.8\% & 86.4\% \\ \hline
$\tau = 10$ & 60.4\% & 64.8\% & 55.6\% & 57.6\% & 51.9\% & 44.5\% & 79.9\% \\ \hline
%start of second table for CIFAR-100
& \multicolumn{7}{c|}{CIFAR-100} \\ \hline
\multicolumn{1}{|l|}{} & \multicolumn{1}{c|}{FGSM} & \multicolumn{1}{c|}{PGD} & \multicolumn{1}{c|}{BPDA} & \multicolumn{1}{c|}{MIM} & \multicolumn{1}{c|}{C\&W} & \multicolumn{1}{c|}{APGD} & \multicolumn{1}{c|}{Acc} \\ \hline
\multicolumn{1}{|l|}{ViT-B-32} & 20.8\% & 1.9\% & 13.4\% & 3.1\% & 0.0\% & 0.0\% & 91.7\% \\ \hline
\multicolumn{1}{|l|}{$\tau = 0$} & 16.2\% & 6.9\% & 9.7\% & 7.6\% & 0.7\% & 1.2\% & 87.6\% \\ \hline
\multicolumn{1}{|l|}{$\tau = 1$} & 21.6\% & 16.5\% & 9.3\% & 12.8\% & 5.2\% & 2.9\% & 81.3\% \\ \hline
\multicolumn{1}{|l|}{$\tau = 2$} & 24.4\% & 23.0\% & 12.3\% & 17.5\% & 10.1\% & 5.4\% & 76.1\% \\ \hline
\multicolumn{1}{|l|}{$\tau = 10$} & 26.4\% & 31.9\% & 20.5\% & 24.3\% & 20.1\% & 18.3\% & 58.6\% \\ \hline
\multicolumn{1}{|l|}{} &  &  &  &  &  &  &  \\ \hline
\multicolumn{1}{|l|}{ViT-B-16} & 20.4\% & 0.0\% & 11.9\% & 0.5\% & 0.0\% & 0.0\% & 92.8\% \\ \hline
\multicolumn{1}{|l|}{$\tau = 0$} & 16.6\% & 8.0\% & 7.6\% & 7.5\% & 0.5\% & 0.3\% & 87.5\% \\ \hline
\multicolumn{1}{|l|}{$\tau = 1$} & 23.9\% & 20.0\% & 9.3\% & 15.7\% & 8.5\% & 4.8\% & 82.0\% \\ \hline
\multicolumn{1}{|l|}{$\tau = 2$} & 26.0\% & 25.0\% & 12.8\% & 18.4\% & 13.3\% & 8.4\% & 77.0\% \\ \hline
\multicolumn{1}{|l|}{$\tau = 10$} & 25.1\% & 29.0\% & 20.4\% & 22.7\% & 16.5\% & 16.1\% & 54.2\% \\ \hline
\multicolumn{1}{|l|}{} & \multicolumn{1}{c|}{} & \multicolumn{1}{c|}{} & \multicolumn{1}{c|}{} & \multicolumn{1}{c|}{} & \multicolumn{1}{c|}{} & \multicolumn{1}{c|}{} & \multicolumn{1}{c|}{} \\ \hline
\multicolumn{1}{|l|}{ViT-L-16} & 33.0\% & 1.6\% & 15.1\% & 4.7\% & 0.0\% & 0.0\% & 94.0\% \\ \hline
\multicolumn{1}{|l|}{$\tau = 0$} & 28.6\% & 19.1\% & 13.1\% & 17.7\% & 5.3\% & 5.2\% & 87.7\% \\ \hline
\multicolumn{1}{|l|}{$\tau = 1$} & 27.7\% & 22.6\% & 14.3\% & 18.1\% & 11.4\% & 6.7\% & 83.0\% \\ \hline
\multicolumn{1}{|l|}{$\tau = 2$} & 30.2\% & 26.5\% & 16.7\% & 22.0\% & 16.0\% & 9.7\% & 80.0\% \\ \hline
\multicolumn{1}{|l|}{$\tau = 10$} & 31.4\% & 32.0\% & 23.0\% & 24.0\% & 20.2\% & 15.3\% & 64.1\% \\ \hline
\multicolumn{1}{|l|}{} &  &  &  &  &  &  &  \\ \hline
\multicolumn{1}{|l|}{ResNet-164} & 7.6\% & 0.3\% & 3.7\% & 0.9\% & 0.0\% & 0.0\% & 74.2\% \\ \hline
\multicolumn{1}{|l|}{$\tau = 0$} & 18.2\% & 16.1\% & 13.5\% & 12.2\% & 9.7\% & 6.8\% & 70.8\% \\ \hline
\multicolumn{1}{|l|}{$\tau = 1$} & 23.5\% & 24.4\% & 19.3\% & 18.4\% & 17.3\% & 10.7\% & 66.8\% \\ \hline
\multicolumn{1}{|l|}{$\tau = 2$} & 35.3\% & 32.3\% & 25.2\% & 26.3\% & 24.7\% & 17.6\% & 61.8\% \\ \hline
\multicolumn{1}{|l|}{$\tau = 10$} & 45.6\% & 43.2\% & 34.5\% & 35.7\% & 28.9\% & 27.2\% & 55.0\% \\ \hline
\end{tabular}
\end{table*}

\section{Additional Tables And Codes}
\label{sup:tables}

In this section we provide full numerical tables for some of the charts and figures presented in our paper. Each table is captioned with a description of which part of the main paper that it corresponds to. 

We also provide code to replicate our results. Code for the ViT/BiT defense, the RayS attack, Adaptive attack and SAGA for CIFAR-10 can be found on Github: \href{https://github.com/MetaMain/ViTRobust}{https://github.com/MetaMain/ViTRobust}.

\begin{table*}
\caption{Full transferability results for CIFAR-10. The first column in each table represents the model used to generate the adversarial examples, $C_{i}$. The top row in each table represents the model used to evaluate the adversarial examples, $C_{j}$. Each entry represents $1-t_{i,j}$ (the robust accuracy) computed using equation~\ref{eq:transfer} with $C_{i}$, $C_{j}$ and either FGSM, PGD or MIM. For PGD and MIM we use only 10 steps to avoid overfitting the example to a paticular model. Based on these results we take the maximum transferability across all attacks and report the result in table~\ref{table:transfer}. We also visually show the maximum transerability $t_{i,j}$ in figure~\ref{fig:transferabilityVisual}.}
{\small
\begin{tabular}{lcccccccc}
 & \multicolumn{8}{c}{FGSM} \\ \cline{2-9} 
\multicolumn{1}{l|}{} & \multicolumn{1}{l|}{ViT-B-32} & \multicolumn{1}{l|}{ViT-B-16} & \multicolumn{1}{l|}{ViT-L-16} & \multicolumn{1}{l|}{R50-ViT-B-16} & \multicolumn{1}{l|}{BiT-M-R50x1} & \multicolumn{1}{l|}{BiT-M-R101x3} & \multicolumn{1}{l|}{ResNet-56} & \multicolumn{1}{l|}{ResNet-164} \\ \hline
\multicolumn{1}{|l|}{ViT-B-32} & \multicolumn{1}{c|}{43.4\%} & \multicolumn{1}{c|}{55.3\%} & \multicolumn{1}{c|}{61.1\%} & \multicolumn{1}{c|}{76.6\%} & \multicolumn{1}{c|}{67.7\%} & \multicolumn{1}{c|}{68.9\%} & \multicolumn{1}{c|}{83.3\%} & \multicolumn{1}{c|}{83.3\%} \\ \hline
\multicolumn{1}{|l|}{ViT-B-16} & \multicolumn{1}{c|}{68.7\%} & \multicolumn{1}{c|}{41.3\%} & \multicolumn{1}{c|}{56.9\%} & \multicolumn{1}{c|}{80.3\%} & \multicolumn{1}{c|}{73.0\%} & \multicolumn{1}{c|}{73.7\%} & \multicolumn{1}{c|}{86.1\%} & \multicolumn{1}{c|}{86.0\%} \\ \hline
\multicolumn{1}{|l|}{ViT-L-16} & \multicolumn{1}{c|}{74.8\%} & \multicolumn{1}{c|}{61.6\%} & \multicolumn{1}{c|}{59.5\%} & \multicolumn{1}{c|}{82.7\%} & \multicolumn{1}{c|}{78.5\%} & \multicolumn{1}{c|}{80.1\%} & \multicolumn{1}{c|}{88.5\%} & \multicolumn{1}{c|}{88.5\%} \\ \hline
\multicolumn{1}{|l|}{R50-ViT-B-16} & \multicolumn{1}{c|}{82.6\%} & \multicolumn{1}{c|}{75.9\%} & \multicolumn{1}{c|}{79.9\%} & \multicolumn{1}{c|}{51.4\%} & \multicolumn{1}{c|}{72.2\%} & \multicolumn{1}{c|}{74.1\%} & \multicolumn{1}{c|}{81.9\%} & \multicolumn{1}{c|}{82.0\%} \\ \hline
\multicolumn{1}{|l|}{BiT-M-R50x1} & \multicolumn{1}{c|}{96.0\%} & \multicolumn{1}{c|}{94.0\%} & \multicolumn{1}{c|}{95.9\%} & \multicolumn{1}{c|}{95.6\%} & \multicolumn{1}{c|}{69.5\%} & \multicolumn{1}{c|}{86.0\%} & \multicolumn{1}{c|}{94.0\%} & \multicolumn{1}{c|}{93.8\%} \\ \hline
\multicolumn{1}{|l|}{BiT-M-R101x3} & \multicolumn{1}{c|}{97.4\%} & \multicolumn{1}{c|}{94.4\%} & \multicolumn{1}{c|}{86.3\%} & \multicolumn{1}{c|}{96.2\%} & \multicolumn{1}{c|}{86.1\%} & \multicolumn{1}{c|}{88.0\%} & \multicolumn{1}{c|}{95.5\%} & \multicolumn{1}{c|}{94.7\%} \\ \hline
\multicolumn{1}{|l|}{ResNet-56} & \multicolumn{1}{c|}{93.4\%} & \multicolumn{1}{c|}{92.4\%} & \multicolumn{1}{c|}{94.7\%} & \multicolumn{1}{c|}{91.2\%} & \multicolumn{1}{c|}{82.0\%} & \multicolumn{1}{c|}{91.2\%} & \multicolumn{1}{c|}{41.9\%} & \multicolumn{1}{c|}{43.0\%} \\ \hline
\multicolumn{1}{|l|}{ResNet-164} & \multicolumn{1}{c|}{93.2\%} & \multicolumn{1}{c|}{92.4\%} & \multicolumn{1}{c|}{95.3\%} & \multicolumn{1}{c|}{90.9\%} & \multicolumn{1}{c|}{82.9\%} & \multicolumn{1}{c|}{91.7\%} & \multicolumn{1}{c|}{45.1\%} & \multicolumn{1}{c|}{47.1\%} \\ \hline
 & \multicolumn{8}{c}{PGD} \\ \cline{2-9} 
\multicolumn{1}{l|}{} & \multicolumn{1}{l|}{ViT-B-32} & \multicolumn{1}{l|}{ViT-B-16} & \multicolumn{1}{l|}{ViT-L-16} & \multicolumn{1}{l|}{R50-ViT-B-16} & \multicolumn{1}{l|}{BiT-M-R50x1} & \multicolumn{1}{l|}{BiT-M-R101x3} & \multicolumn{1}{l|}{ResNet-56} & \multicolumn{1}{l|}{ResNet-164} \\ \hline
\multicolumn{1}{|l|}{ViT-B-32} & \multicolumn{1}{c|}{4.2\%} & \multicolumn{1}{c|}{49.1\%} & \multicolumn{1}{c|}{72.2\%} & \multicolumn{1}{c|}{93.4\%} & \multicolumn{1}{c|}{70.4\%} & \multicolumn{1}{c|}{74.3\%} & \multicolumn{1}{c|}{93.0\%} & \multicolumn{1}{c|}{92.9\%} \\ \hline
\multicolumn{1}{|l|}{ViT-B-16} & \multicolumn{1}{c|}{90.4\%} & \multicolumn{1}{c|}{0.4\%} & \multicolumn{1}{c|}{53.9\%} & \multicolumn{1}{c|}{97.1\%} & \multicolumn{1}{c|}{83.3\%} & \multicolumn{1}{c|}{85.1\%} & \multicolumn{1}{c|}{96.1\%} & \multicolumn{1}{c|}{95.9\%} \\ \hline
\multicolumn{1}{|l|}{ViT-L-16} & \multicolumn{1}{c|}{85.9\%} & \multicolumn{1}{c|}{32.4\%} & \multicolumn{1}{c|}{10.4\%} & \multicolumn{1}{c|}{94.4\%} & \multicolumn{1}{c|}{82.1\%} & \multicolumn{1}{c|}{81.7\%} & \multicolumn{1}{c|}{95.7\%} & \multicolumn{1}{c|}{95.5\%} \\ \hline
\multicolumn{1}{|l|}{R50-ViT-B-16} & \multicolumn{1}{c|}{93.7\%} & \multicolumn{1}{c|}{84.5\%} & \multicolumn{1}{c|}{91.3\%} & \multicolumn{1}{c|}{6.6\%} & \multicolumn{1}{c|}{69.7\%} & \multicolumn{1}{c|}{75.3\%} & \multicolumn{1}{c|}{90.3\%} & \multicolumn{1}{c|}{88.9\%} \\ \hline
\multicolumn{1}{|l|}{BiT-M-R101x3} & \multicolumn{1}{c|}{99.6\%} & \multicolumn{1}{c|}{97.5\%} & \multicolumn{1}{c|}{86.3\%} & \multicolumn{1}{c|}{98.7\%} & \multicolumn{1}{c|}{58.0\%} & \multicolumn{1}{c|}{0.0\%} & \multicolumn{1}{c|}{97.3\%} & \multicolumn{1}{c|}{96.8\%} \\ \hline
\multicolumn{1}{|l|}{BiT-M-R50x1} & \multicolumn{1}{c|}{99.9\%} & \multicolumn{1}{c|}{98.5\%} & \multicolumn{1}{c|}{99.4\%} & \multicolumn{1}{c|}{99.5\%} & \multicolumn{1}{c|}{0.0\%} & \multicolumn{1}{c|}{85.9\%} & \multicolumn{1}{c|}{98.2\%} & \multicolumn{1}{c|}{97.8\%} \\ \hline
\multicolumn{1}{|l|}{ResNet-56} & \multicolumn{1}{c|}{99.0\%} & \multicolumn{1}{c|}{97.9\%} & \multicolumn{1}{c|}{98.6\%} & \multicolumn{1}{c|}{98.4\%} & \multicolumn{1}{c|}{93.9\%} & \multicolumn{1}{c|}{96.9\%} & \multicolumn{1}{c|}{28.0\%} & \multicolumn{1}{c|}{28.7\%} \\ \hline
\multicolumn{1}{|l|}{ResNet-164} & \multicolumn{1}{c|}{98.7\%} & \multicolumn{1}{c|}{98.3\%} & \multicolumn{1}{c|}{99.1\%} & \multicolumn{1}{c|}{98.1\%} & \multicolumn{1}{c|}{92.6\%} & \multicolumn{1}{c|}{96.1\%} & \multicolumn{1}{c|}{28.7\%} & \multicolumn{1}{c|}{32.5\%} \\ \hline
 & \multicolumn{8}{c}{MIM} \\ \cline{2-9} 
\multicolumn{1}{l|}{} & \multicolumn{1}{l|}{ViT-B-32} & \multicolumn{1}{l|}{ViT-B-16} & \multicolumn{1}{l|}{ViT-L-16} & \multicolumn{1}{l|}{R50-ViT-B-16} & \multicolumn{1}{l|}{BiT-M-R50x1} & \multicolumn{1}{l|}{BiT-M-R101x3} & \multicolumn{1}{l|}{ResNet-56} & \multicolumn{1}{l|}{ResNet-164} \\ \hline
\multicolumn{1}{|l|}{ViT-B-32} & \multicolumn{1}{c|}{4.9\%} & \multicolumn{1}{c|}{15.9\%} & \multicolumn{1}{c|}{24.5\%} & \multicolumn{1}{c|}{65.1\%} & \multicolumn{1}{c|}{39.2\%} & \multicolumn{1}{c|}{38.0\%} & \multicolumn{1}{c|}{81.4\%} & \multicolumn{1}{c|}{80.1\%} \\ \hline
\multicolumn{1}{|l|}{ViT-B-16} & \multicolumn{1}{c|}{42.9\%} & \multicolumn{1}{c|}{0.9\%} & \multicolumn{1}{c|}{11.1\%} & \multicolumn{1}{c|}{77.4\%} & \multicolumn{1}{c|}{56.6\%} & \multicolumn{1}{c|}{55.0\%} & \multicolumn{1}{c|}{86.9\%} & \multicolumn{1}{c|}{86.0\%} \\ \hline
\multicolumn{1}{|l|}{ViT-L-16} & \multicolumn{1}{c|}{44.4\%} & \multicolumn{1}{c|}{21.6\%} & \multicolumn{1}{c|}{13.4\%} & \multicolumn{1}{c|}{69.7\%} & \multicolumn{1}{c|}{57.5\%} & \multicolumn{1}{c|}{55.3\%} & \multicolumn{1}{c|}{87.0\%} & \multicolumn{1}{c|}{85.2\%} \\ \hline
\multicolumn{1}{|l|}{R50-ViT-B-16} & \multicolumn{1}{c|}{60.4\%} & \multicolumn{1}{c|}{41.9\%} & \multicolumn{1}{c|}{48.5\%} & \multicolumn{1}{c|}{1.7\%} & \multicolumn{1}{c|}{39.0\%} & \multicolumn{1}{c|}{42.0\%} & \multicolumn{1}{c|}{73.3\%} & \multicolumn{1}{c|}{71.0\%} \\ \hline
\multicolumn{1}{|l|}{BiT-M-R50x1} & \multicolumn{1}{c|}{95.5\%} & \multicolumn{1}{c|}{89.1\%} & \multicolumn{1}{c|}{94.3\%} & \multicolumn{1}{c|}{95.3\%} & \multicolumn{1}{c|}{0.0\%} & \multicolumn{1}{c|}{48.6\%} & \multicolumn{1}{c|}{93.0\%} & \multicolumn{1}{c|}{91.0\%} \\ \hline
\multicolumn{1}{|l|}{BiT-M-R101x3} & \multicolumn{1}{c|}{91.4\%} & \multicolumn{1}{c|}{79.7\%} & \multicolumn{1}{c|}{88.0\%} & \multicolumn{1}{c|}{92.8\%} & \multicolumn{1}{c|}{24.1\%} & \multicolumn{1}{c|}{0.1\%} & \multicolumn{1}{c|}{92.2\%} & \multicolumn{1}{c|}{90.7\%} \\ \hline
\multicolumn{1}{|l|}{ResNet-56} & \multicolumn{1}{c|}{94.2\%} & \multicolumn{1}{c|}{91.0\%} & \multicolumn{1}{c|}{94.8\%} & \multicolumn{1}{c|}{90.3\%} & \multicolumn{1}{c|}{77.5\%} & \multicolumn{1}{c|}{88.2\%} & \multicolumn{1}{c|}{14.1\%} & \multicolumn{1}{c|}{12.8\%} \\ \hline
\multicolumn{1}{|l|}{ResNet-164} & \multicolumn{1}{c|}{94.2\%} & \multicolumn{1}{c|}{91.9\%} & \multicolumn{1}{c|}{95.0\%} & \multicolumn{1}{c|}{90.3\%} & \multicolumn{1}{c|}{77.7\%} & \multicolumn{1}{c|}{88.8\%} & \multicolumn{1}{c|}{16.4\%} & \multicolumn{1}{c|}{14.3\%} \\ \hline
\end{tabular}
}
\end{table*}

\begin{table*}[]
\caption{Full transferability results for CIFAR-100. The first column in each table represents the model used to generate the adversarial examples, $C_{i}$. The top row in each table represents the model used to evaluate the adversarial examples, $C_{j}$. Each entry represents $1-t_{i,j}$ (the robust accuracy) computed using equation~\ref{eq:transfer} with $C_{i}$, $C_{j}$ and either FGSM, PGD or MIM. For PGD and MIM we use only 10 steps to avoid overfitting the example to a paticular model. Based on these results we take the maximum transferability across all attacks and report the result in table~\ref{table:transfer}.}
{\small
\begin{tabular}{lcccccccc}
 & \multicolumn{8}{c}{FGSM} \\ \cline{2-9} 
\multicolumn{1}{l|}{} & \multicolumn{1}{l|}{ViT-B-32} & \multicolumn{1}{l|}{ViT-B-16} & \multicolumn{1}{l|}{ViT-L-16} & \multicolumn{1}{l|}{R50-ViT-B-16} & \multicolumn{1}{l|}{BiT-M-R50x1} & \multicolumn{1}{l|}{BiT-M-R101x3} & \multicolumn{1}{l|}{ResNet-56} & \multicolumn{1}{l|}{ResNet-164} \\ \hline
\multicolumn{1}{|l|}{ViT-B-32} & \multicolumn{1}{c|}{27.9\%} & \multicolumn{1}{c|}{40.2\%} & \multicolumn{1}{c|}{41.7\%} & \multicolumn{1}{c|}{59.2\%} & \multicolumn{1}{c|}{55.8\%} & \multicolumn{1}{c|}{57.4\%} & \multicolumn{1}{c|}{85.1\%} & \multicolumn{1}{c|}{86.0\%} \\ \hline
\multicolumn{1}{|l|}{ViT-B-16} & \multicolumn{1}{c|}{50.7\%} & \multicolumn{1}{c|}{25.3\%} & \multicolumn{1}{c|}{36.7\%} & \multicolumn{1}{c|}{64.9\%} & \multicolumn{1}{c|}{61.4\%} & \multicolumn{1}{c|}{59.9\%} & \multicolumn{1}{c|}{91.0\%} & \multicolumn{1}{c|}{92.5\%} \\ \hline
\multicolumn{1}{|l|}{ViT-L-16} & \multicolumn{1}{c|}{57.0\%} & \multicolumn{1}{c|}{41.7\%} & \multicolumn{1}{c|}{37.8\%} & \multicolumn{1}{c|}{65.7\%} & \multicolumn{1}{c|}{64.7\%} & \multicolumn{1}{c|}{65.1\%} & \multicolumn{1}{c|}{90.1\%} & \multicolumn{1}{c|}{90.5\%} \\ \hline
\multicolumn{1}{|l|}{R50-ViT-B-16} & \multicolumn{1}{c|}{66.3\%} & \multicolumn{1}{c|}{60.6\%} & \multicolumn{1}{c|}{64.3\%} & \multicolumn{1}{c|}{30.8\%} & \multicolumn{1}{c|}{56.1\%} & \multicolumn{1}{c|}{61.2\%} & \multicolumn{1}{c|}{90.8\%} & \multicolumn{1}{c|}{91.4\%} \\ \hline
\multicolumn{1}{|l|}{BiT-M-R50x1} & \multicolumn{1}{c|}{87.3\%} & \multicolumn{1}{c|}{83.1\%} & \multicolumn{1}{c|}{87.2\%} & \multicolumn{1}{c|}{86.0\%} & \multicolumn{1}{c|}{44.5\%} & \multicolumn{1}{c|}{68.8\%} & \multicolumn{1}{c|}{96.3\%} & \multicolumn{1}{c|}{96.6\%} \\ \hline
\multicolumn{1}{|l|}{BiT-M-R101x3} & \multicolumn{1}{c|}{85.5\%} & \multicolumn{1}{c|}{83.3\%} & \multicolumn{1}{c|}{85.8\%} & \multicolumn{1}{c|}{86.4\%} & \multicolumn{1}{c|}{70.4\%} & \multicolumn{1}{c|}{67.0\%} & \multicolumn{1}{c|}{96.2\%} & \multicolumn{1}{c|}{97.8\%} \\ \hline
\multicolumn{1}{|l|}{ResNet-56} & \multicolumn{1}{c|}{79.9\%} & \multicolumn{1}{c|}{77.8\%} & \multicolumn{1}{c|}{84.7\%} & \multicolumn{1}{c|}{77.3\%} & \multicolumn{1}{c|}{68.6\%} & \multicolumn{1}{c|}{78.1\%} & \multicolumn{1}{c|}{38.0\%} & \multicolumn{1}{c|}{40.8\%} \\ \hline
\multicolumn{1}{|l|}{ResNet-164} & \multicolumn{1}{c|}{77.9\%} & \multicolumn{1}{c|}{75.5\%} & \multicolumn{1}{c|}{84.5\%} & \multicolumn{1}{c|}{75.8\%} & \multicolumn{1}{c|}{64.9\%} & \multicolumn{1}{c|}{75.7\%} & \multicolumn{1}{c|}{36.1\%} & \multicolumn{1}{c|}{33.2\%} \\ \hline
 & \multicolumn{8}{c}{PGD} \\ \cline{2-9} 
\multicolumn{1}{l|}{} & \multicolumn{1}{l|}{ViT-B-32} & \multicolumn{1}{l|}{ViT-B-16} & \multicolumn{1}{l|}{ViT-L-16} & \multicolumn{1}{l|}{R50-ViT-B-16} & \multicolumn{1}{l|}{BiT-M-R50x1} & \multicolumn{1}{l|}{BiT-M-R101x3} & \multicolumn{1}{l|}{ResNet-56} & \multicolumn{1}{l|}{ResNet-164} \\ \hline
\multicolumn{1}{|l|}{ViT-B-32} & \multicolumn{1}{c|}{3.8\%} & \multicolumn{1}{c|}{36.4\%} & \multicolumn{1}{c|}{53.5\%} & \multicolumn{1}{c|}{80.9\%} & \multicolumn{1}{c|}{67.7\%} & \multicolumn{1}{c|}{68.5\%} & \multicolumn{1}{c|}{93.3\%} & \multicolumn{1}{c|}{95.0\%} \\ \hline
\multicolumn{1}{|l|}{ViT-B-16} & \multicolumn{1}{c|}{78.6\%} & \multicolumn{1}{c|}{0.7\%} & \multicolumn{1}{c|}{36.1\%} & \multicolumn{1}{c|}{90.3\%} & \multicolumn{1}{c|}{78.9\%} & \multicolumn{1}{c|}{80.2\%} & \multicolumn{1}{c|}{97.2\%} & \multicolumn{1}{c|}{97.6\%} \\ \hline
\multicolumn{1}{|l|}{ViT-L-16} & \multicolumn{1}{c|}{72.2\%} & \multicolumn{1}{c|}{17.9\%} & \multicolumn{1}{c|}{5.8\%} & \multicolumn{1}{c|}{85.0\%} & \multicolumn{1}{c|}{76.6\%} & \multicolumn{1}{c|}{75.2\%} & \multicolumn{1}{c|}{96.1\%} & \multicolumn{1}{c|}{96.2\%} \\ \hline
\multicolumn{1}{|l|}{R50-ViT-B-16} & \multicolumn{1}{c|}{85.6\%} & \multicolumn{1}{c|}{75.5\%} & \multicolumn{1}{c|}{82.3\%} & \multicolumn{1}{c|}{2.2\%} & \multicolumn{1}{c|}{62.7\%} & \multicolumn{1}{c|}{68.5\%} & \multicolumn{1}{c|}{94.3\%} & \multicolumn{1}{c|}{95.3\%} \\ \hline
\multicolumn{1}{|l|}{BiT-M-R50x1} & \multicolumn{1}{c|}{96.1\%} & \multicolumn{1}{c|}{94.7\%} & \multicolumn{1}{c|}{97.8\%} & \multicolumn{1}{c|}{96.1\%} & \multicolumn{1}{c|}{0.0\%} & \multicolumn{1}{c|}{76.9\%} & \multicolumn{1}{c|}{97.5\%} & \multicolumn{1}{c|}{98.7\%} \\ \hline
\multicolumn{1}{|l|}{BiT-M-R101x3} & \multicolumn{1}{c|}{94.8\%} & \multicolumn{1}{c|}{91.7\%} & \multicolumn{1}{c|}{95.2\%} & \multicolumn{1}{c|}{94.1\%} & \multicolumn{1}{c|}{52.3\%} & \multicolumn{1}{c|}{1.0\%} & \multicolumn{1}{c|}{97.9\%} & \multicolumn{1}{c|}{97.9\%} \\ \hline
\multicolumn{1}{|l|}{ResNet-56} & \multicolumn{1}{c|}{91.6\%} & \multicolumn{1}{c|}{91.0\%} & \multicolumn{1}{c|}{94.5\%} & \multicolumn{1}{c|}{89.4\%} & \multicolumn{1}{c|}{82.0\%} & \multicolumn{1}{c|}{89.9\%} & \multicolumn{1}{c|}{51.2\%} & \multicolumn{1}{c|}{56.6\%} \\ \hline
\multicolumn{1}{|l|}{ResNet-164} & \multicolumn{1}{c|}{89.2\%} & \multicolumn{1}{c|}{89.0\%} & \multicolumn{1}{c|}{92.8\%} & \multicolumn{1}{c|}{88.5\%} & \multicolumn{1}{c|}{78.4\%} & \multicolumn{1}{c|}{85.7\%} & \multicolumn{1}{c|}{43.6\%} & \multicolumn{1}{c|}{39.4\%} \\ \hline
 & \multicolumn{8}{c}{MIM} \\ \cline{2-9} 
\multicolumn{1}{l|}{} & \multicolumn{1}{l|}{ViT-B-32} & \multicolumn{1}{l|}{ViT-B-16} & \multicolumn{1}{l|}{ViT-L-16} & \multicolumn{1}{l|}{R50-ViT-B-16} & \multicolumn{1}{l|}{BiT-M-R50x1} & \multicolumn{1}{l|}{BiT-M-R101x3} & \multicolumn{1}{l|}{ResNet-56} & \multicolumn{1}{l|}{ResNet-164} \\ \hline
\multicolumn{1}{|l|}{ViT-B-32} & \multicolumn{1}{c|}{4.4\%} & \multicolumn{1}{c|}{11.5\%} & \multicolumn{1}{c|}{16.4\%} & \multicolumn{1}{c|}{47.8\%} & \multicolumn{1}{c|}{39.5\%} & \multicolumn{1}{c|}{38.9\%} & \multicolumn{1}{c|}{86.1\%} & \multicolumn{1}{c|}{87.4\%} \\ \hline
\multicolumn{1}{|l|}{ViT-B-16} & \multicolumn{1}{c|}{28.7\%} & \multicolumn{1}{c|}{0.9\%} & \multicolumn{1}{c|}{6.8\%} & \multicolumn{1}{c|}{61.4\%} & \multicolumn{1}{c|}{55.5\%} & \multicolumn{1}{c|}{52.1\%} & \multicolumn{1}{c|}{91.4\%} & \multicolumn{1}{c|}{93.3\%} \\ \hline
\multicolumn{1}{|l|}{ViT-L-16} & \multicolumn{1}{c|}{32.2\%} & \multicolumn{1}{c|}{11.7\%} & \multicolumn{1}{c|}{7.5\%} & \multicolumn{1}{c|}{51.9\%} & \multicolumn{1}{c|}{52.4\%} & \multicolumn{1}{c|}{50.0\%} & \multicolumn{1}{c|}{91.3\%} & \multicolumn{1}{c|}{90.5\%} \\ \hline
\multicolumn{1}{|l|}{R50-ViT-B-16} & \multicolumn{1}{c|}{48.4\%} & \multicolumn{1}{c|}{35.0\%} & \multicolumn{1}{c|}{37.7\%} & \multicolumn{1}{c|}{1.1\%} & \multicolumn{1}{c|}{35.9\%} & \multicolumn{1}{c|}{38.8\%} & \multicolumn{1}{c|}{89.0\%} & \multicolumn{1}{c|}{90.1\%} \\ \hline
\multicolumn{1}{|l|}{BiT-M-R50x1} & \multicolumn{1}{c|}{82.3\%} & \multicolumn{1}{c|}{75.0\%} & \multicolumn{1}{c|}{84.5\%} & \multicolumn{1}{c|}{81.8\%} & \multicolumn{1}{c|}{0.1\%} & \multicolumn{1}{c|}{43.5\%} & \multicolumn{1}{c|}{95.1\%} & \multicolumn{1}{c|}{94.8\%} \\ \hline
\multicolumn{1}{|l|}{BiT-M-R101x3} & \multicolumn{1}{c|}{75.1\%} & \multicolumn{1}{c|}{61.0\%} & \multicolumn{1}{c|}{73.7\%} & \multicolumn{1}{c|}{76.5\%} & \multicolumn{1}{c|}{26.0\%} & \multicolumn{1}{c|}{1.2\%} & \multicolumn{1}{c|}{94.3\%} & \multicolumn{1}{c|}{96.8\%} \\ \hline
\multicolumn{1}{|l|}{ResNet-56} & \multicolumn{1}{c|}{81.4\%} & \multicolumn{1}{c|}{80.5\%} & \multicolumn{1}{c|}{86.4\%} & \multicolumn{1}{c|}{78.9\%} & \multicolumn{1}{c|}{69.3\%} & \multicolumn{1}{c|}{79.5\%} & \multicolumn{1}{c|}{29.2\%} & \multicolumn{1}{c|}{31.1\%} \\ \hline
\multicolumn{1}{|l|}{ResNet-164} & \multicolumn{1}{c|}{78.3\%} & \multicolumn{1}{c|}{77.2\%} & \multicolumn{1}{c|}{84.8\%} & \multicolumn{1}{c|}{76.5\%} & \multicolumn{1}{c|}{64.1\%} & \multicolumn{1}{c|}{73.5\%} & \multicolumn{1}{c|}{25.5\%} & \multicolumn{1}{c|}{20.8\%} \\ \hline
\end{tabular}
}
\end{table*}

\begin{table*}[]
\caption{Full transferability results for ImageNet. The first column in each table represents the model used to generate the adversarial examples, $C_{i}$. The top row in each table represents the model used to evaluate the adversarial examples, $C_{j}$. Each entry represents $1-t_{i,j}$ (the robust accuracy) computed using equation~\ref{eq:transfer} with $C_{i}$, $C_{j}$ and either FGSM, PGD or MIM. For PGD and MIM we use only 10 steps to avoid overfitting the example to a paticular model. Based on these results we take the maximum transferability across all attacks and report the result in table~\ref{table:transfer}. Note due to the complexity of training ImageNet models we do not train independent copies of the model to measure self-transferability (when $i=j$).}
{\small
\begin{tabular}{lccccccc}
 & \multicolumn{7}{c}{FGSM} \\ \cline{2-8} 
\multicolumn{1}{l|}{} & \multicolumn{1}{l|}{ViT-B-16} & \multicolumn{1}{l|}{ViT-L-16 (224)} & \multicolumn{1}{l|}{ViT-L-16 (512)} & \multicolumn{1}{l|}{BiT-M-R50x1} & \multicolumn{1}{l|}{BiT-M-R152x4} & \multicolumn{1}{l|}{ResNet-50} & \multicolumn{1}{l|}{ResNet-152} \\ \hline
\multicolumn{1}{|l|}{ViT-B-16} & \multicolumn{1}{c|}{+} & \multicolumn{1}{c|}{40.8\%} & \multicolumn{1}{c|}{67.3\%} & \multicolumn{1}{c|}{63.2\%} & \multicolumn{1}{c|}{73.2\%} & \multicolumn{1}{c|}{56.0\%} & \multicolumn{1}{c|}{63.6\%} \\ \hline
\multicolumn{1}{|l|}{ViT-L-16 (224)} & \multicolumn{1}{c|}{40.1\%} & \multicolumn{1}{c|}{+} & \multicolumn{1}{c|}{59.6\%} & \multicolumn{1}{c|}{63.7\%} & \multicolumn{1}{c|}{75.4\%} & \multicolumn{1}{c|}{57.6\%} & \multicolumn{1}{c|}{61.7\%} \\ \hline
\multicolumn{1}{|l|}{ViT-L-16 (512)} & \multicolumn{1}{c|}{77.8\%} & \multicolumn{1}{c|}{69.3\%} & \multicolumn{1}{c|}{+} & \multicolumn{1}{c|}{74.6\%} & \multicolumn{1}{c|}{77.7\%} & \multicolumn{1}{c|}{74.5\%} & \multicolumn{1}{c|}{78.4\%} \\ \hline
\multicolumn{1}{|l|}{BiT-M-R50x1} & \multicolumn{1}{c|}{90.6\%} & \multicolumn{1}{c|}{91.6\%} & \multicolumn{1}{c|}{89.4\%} & \multicolumn{1}{c|}{+} & \multicolumn{1}{c|}{83.3\%} & \multicolumn{1}{c|}{81.2\%} & \multicolumn{1}{c|}{83.5\%} \\ \hline
\multicolumn{1}{|l|}{BiT-M-R152x4} & \multicolumn{1}{c|}{93.0\%} & \multicolumn{1}{c|}{93.9\%} & \multicolumn{1}{c|}{89.8\%} & \multicolumn{1}{c|}{83.4\%} & \multicolumn{1}{c|}{+} & \multicolumn{1}{c|}{86.8\%} & \multicolumn{1}{c|}{90.1\%} \\ \hline
\multicolumn{1}{|l|}{ResNet-50} & \multicolumn{1}{c|}{77.8\%} & \multicolumn{1}{c|}{82.3\%} & \multicolumn{1}{c|}{79.1\%} & \multicolumn{1}{c|}{61.6\%} & \multicolumn{1}{c|}{79.2\%} & \multicolumn{1}{c|}{+} & \multicolumn{1}{c|}{46.7\%} \\ \hline
\multicolumn{1}{|l|}{ResNet-152} & \multicolumn{1}{c|}{75.6\%} & \multicolumn{1}{c|}{78.8\%} & \multicolumn{1}{c|}{77.9\%} & \multicolumn{1}{c|}{61.0\%} & \multicolumn{1}{c|}{78.1\%} & \multicolumn{1}{c|}{40.7\%} & \multicolumn{1}{c|}{+} \\ \hline
 & \multicolumn{7}{c}{PGD} \\ \cline{2-8} 
\multicolumn{1}{l|}{} & \multicolumn{1}{l|}{ViT-B-16} & \multicolumn{1}{l|}{ViT-L-16 (224)} & \multicolumn{1}{l|}{ViT-L-16 (512)} & \multicolumn{1}{l|}{BiT-M-R50x1} & \multicolumn{1}{l|}{BiT-M-R152x4} & \multicolumn{1}{l|}{ResNet-50} & \multicolumn{1}{l|}{ResNet-152} \\ \hline
\multicolumn{1}{|l|}{ViT-B-16} & \multicolumn{1}{c|}{+} & \multicolumn{1}{c|}{36.1\%} & \multicolumn{1}{c|}{81.1\%} & \multicolumn{1}{c|}{83.1\%} & \multicolumn{1}{c|}{89.7\%} & \multicolumn{1}{c|}{79.0\%} & \multicolumn{1}{c|}{81.7\%} \\ \hline
\multicolumn{1}{|l|}{ViT-L-16 (224)} & \multicolumn{1}{c|}{22.7\%} & \multicolumn{1}{c|}{+} & \multicolumn{1}{c|}{62.6\%} & \multicolumn{1}{c|}{83.2\%} & \multicolumn{1}{c|}{88.8\%} & \multicolumn{1}{c|}{80.4\%} & \multicolumn{1}{c|}{80.6\%} \\ \hline
\multicolumn{1}{|l|}{ViT-L-16 (512)} & \multicolumn{1}{c|}{89.6\%} & \multicolumn{1}{c|}{83.5\%} & \multicolumn{1}{c|}{+} & \multicolumn{1}{c|}{84.3\%} & \multicolumn{1}{c|}{87.6\%} & \multicolumn{1}{c|}{87.6\%} & \multicolumn{1}{c|}{89.9\%} \\ \hline
\multicolumn{1}{|l|}{BiT-M-R50x1} & \multicolumn{1}{c|}{96.5\%} & \multicolumn{1}{c|}{96.8\%} & \multicolumn{1}{c|}{95.8\%} & \multicolumn{1}{c|}{+} & \multicolumn{1}{c|}{90.6\%} & \multicolumn{1}{c|}{89.2\%} & \multicolumn{1}{c|}{91.4\%} \\ \hline
\multicolumn{1}{|l|}{BiT-M-R152x4} & \multicolumn{1}{c|}{91.8\%} & \multicolumn{1}{c|}{97.3\%} & \multicolumn{1}{c|}{94.2\%} & \multicolumn{1}{c|}{85.4\%} & \multicolumn{1}{c|}{+} & \multicolumn{1}{c|}{93.0\%} & \multicolumn{1}{c|}{95.2\%} \\ \hline
\multicolumn{1}{|l|}{ResNet-50} & \multicolumn{1}{c|}{92.7\%} & \multicolumn{1}{c|}{94.2\%} & \multicolumn{1}{c|}{91.8\%} & \multicolumn{1}{c|}{77.8\%} & \multicolumn{1}{c|}{92.7\%} & \multicolumn{1}{c|}{+} & \multicolumn{1}{c|}{42.2\%} \\ \hline
\multicolumn{1}{|l|}{ResNet-152} & \multicolumn{1}{c|}{91.1\%} & \multicolumn{1}{c|}{93.3\%} & \multicolumn{1}{c|}{90.5\%} & \multicolumn{1}{c|}{77.4\%} & \multicolumn{1}{c|}{90.9\%} & \multicolumn{1}{c|}{30.1\%} & \multicolumn{1}{c|}{+} \\ \hline
 & \multicolumn{7}{c}{MIM} \\ \cline{2-8} 
\multicolumn{1}{l|}{} & \multicolumn{1}{l|}{ViT-B-16} & \multicolumn{1}{l|}{ViT-L-16 (224)} & \multicolumn{1}{l|}{ViT-L-16 (512)} & \multicolumn{1}{l|}{BiT-M-R50x1} & \multicolumn{1}{l|}{BiT-M-R152x4} & \multicolumn{1}{l|}{ResNet-50} & \multicolumn{1}{l|}{ResNet-152} \\ \hline
\multicolumn{1}{|l|}{ViT-B-16} & \multicolumn{1}{c|}{+} & \multicolumn{1}{c|}{10.9\%} & \multicolumn{1}{c|}{60.4\%} & \multicolumn{1}{c|}{59.2\%} & \multicolumn{1}{c|}{72.6\%} & \multicolumn{1}{c|}{56.6\%} & \multicolumn{1}{c|}{59.9\%} \\ \hline
\multicolumn{1}{|l|}{ViT-L-16 (224)} & \multicolumn{1}{c|}{9.1\%} & \multicolumn{1}{c|}{+} & \multicolumn{1}{c|}{35.5\%} & \multicolumn{1}{c|}{60.0\%} & \multicolumn{1}{c|}{73.1\%} & \multicolumn{1}{c|}{56.3\%} & \multicolumn{1}{c|}{59.2\%} \\ \hline
\multicolumn{1}{|l|}{ViT-L-16 (512)} & \multicolumn{1}{c|}{72.0\%} & \multicolumn{1}{c|}{56.6\%} & \multicolumn{1}{c|}{+} & \multicolumn{1}{c|}{65.7\%} & \multicolumn{1}{c|}{73.7\%} & \multicolumn{1}{c|}{71.6\%} & \multicolumn{1}{c|}{76.8\%} \\ \hline
\multicolumn{1}{|l|}{BiT-M-R50x1} & \multicolumn{1}{c|}{90.2\%} & \multicolumn{1}{c|}{91.6\%} & \multicolumn{1}{c|}{88.2\%} & \multicolumn{1}{c|}{+} & \multicolumn{1}{c|}{75.1\%} & \multicolumn{1}{c|}{75.3\%} & \multicolumn{1}{c|}{81.3\%} \\ \hline
\multicolumn{1}{|l|}{BiT-M-R152x4} & \multicolumn{1}{c|}{96.2\%} & \multicolumn{1}{c|}{92.4\%} & \multicolumn{1}{c|}{86.5\%} & \multicolumn{1}{c|}{72.0\%} & \multicolumn{1}{c|}{+} & \multicolumn{1}{c|}{84.9\%} & \multicolumn{1}{c|}{88.0\%} \\ \hline
\multicolumn{1}{|l|}{ResNet-50} & \multicolumn{1}{c|}{76.2\%} & \multicolumn{1}{c|}{81.2\%} & \multicolumn{1}{c|}{75.3\%} & \multicolumn{1}{c|}{44.7\%} & \multicolumn{1}{c|}{75.6\%} & \multicolumn{1}{c|}{+} & \multicolumn{1}{c|}{13.3\%} \\ \hline
\multicolumn{1}{|l|}{ResNet-152} & \multicolumn{1}{c|}{74.1\%} & \multicolumn{1}{c|}{77.9\%} & \multicolumn{1}{c|}{73.4\%} & \multicolumn{1}{c|}{45.9\%} & \multicolumn{1}{c|}{73.2\%} & \multicolumn{1}{c|}{10.6\%} & \multicolumn{1}{c|}{+} \\ \hline
\end{tabular}
}
\end{table*}
\end{document}